\documentclass[journal,twoside,web]{ieeecolor}
\usepackage{generic}
\usepackage{amsmath,amssymb,amsfonts}
\usepackage{dsfont}
\usepackage{bm}
\usepackage{algorithmic}
\usepackage{acronym}
\usepackage{graphicx}
\usepackage{textcomp}
\usepackage{stackengine}
\usepackage[table]{xcolor}
\usepackage{tikz}
\usepackage{pgf}

\makeatletter
\let\NAT@parse\undefined
\makeatother
\usepackage[hidelinks]{hyperref}
\usepackage[capitalise,noabbrev]{cleveref}
\usepackage{cite}

\definecolor{diag}{gray}{0.5}
\definecolor{site}{RGB}{200,255,200}
\definecolor{morp}{RGB}{255,200,200}

\DeclareMathOperator*{\argmax}{arg\,max}

\newcommand{\vect}[1]{\bm{#1}}

\def\loss{\mathcal{L}}

\def\aaa{\phantom{$^{***}$}}
\def\aa{\phantom{$^{**}$}}
\def\a{\phantom{$^{*}$}}

\def\ONE#1{\mathds{1}\!\left\{#1\right\}}

\newcommand{\maxp}{\textbf{\emph{MAX}}}
\newcommand{\softmax}{\textbf{\emph{ATT}}}
\newcommand{\maxi}{\textbf{\emph{MAXi}}}

\newcommand{\maxh}{\textbf{\emph{MAXh}}}
\newcommand{\softmaxh}{\textbf{\emph{ATTh}}}
\newcommand{\bert}{\textbf{\emph{BERT}}}
\newcommand{\cnn}{\textbf{\emph{CNN}}}
\newcommand{\gru}{\textbf{\emph{GRU}}}
\newcommand{\svm}{\textbf{\emph{SVM}}}

\acrodef{max}[MM]{Max Model}
\acrodef{softmax}[AM]{Attention Model}
\acrodef{maxi}[MMi]{Max Model interpretable}
\acrodef{maxh}[MMh]{Max Model hierarchical}
\acrodef{softmaxh}[AMh]{Attention Model hierarchical}
\acrodef{rtt}[RTT]{Registro Tumori della Toscana, Tumor Register of Tuscany}
\acrodef{icdo}[ICD-O]{International Classification of Diseases for Oncology}
\acrodef{icdo1}[ICD-O1]{International Classification of Diseases for Oncology, first edition}
\acrodef{icdo3}[ICD-O3]{International Classification of Diseases for Oncology, third edition}
\acrodef{hdr}[HDR]{Hospital Discharge Register}
\acrodef{cdf}[CDF]{Cumulative Distribution Function}
\acrodef{an}[AN]{Artificial Neuron}
\acrodef{ann}[ANN]{Artificial Neural Network}
\acrodef{cnn}[CNN]{Convolutional Neural Network}
\acrodef{rnn}[RNN]{Recurrent Neural Network}
\acrodef{mlp}[MLP]{Multilayer Perceptron}
\acrodef{lstm}[LSTM]{Long Short-Term Memory}
\acrodef{sgd}[SGD]{Stochastic Gradient Descend}
\acrodef{glove}[GloVe]{Global Vectors}
\acrodef{nb}[NB]{Naive Bayes}
\acrodef{svm}[SVM]{Support Vector Machine}
\acrodef{tfidf}[TF-IDF]{Term-Frequency Inverse-Document-Frequency}
\acrodef{map}[MAP]{Mean Average Precision}
\acrodef{relu}[ReLU]{Rectified Linear Unit}
\acrodef{gru}[GRU]{Gated Recurrent Unit}
\acrodef{nlp}[NLP]{Natural Language Processing}
\acrodef{bert}[BERT]{Bidirectional Encoder Representations from Transformers}
\acrodef{mlm}[MLM]{Masked Language Model}
\acrodef{nsp}[NSP]{Next Sentence Prediction}
\acrodef{lm}[LM]{Language Model}

\newcommand\attTableIcdoWidth{1.85cm}
\newcommand\attTableTextWidth{6.4cm}
\newcommand\attHigh{100}
\newcommand\attMid{50}
\newcommand\attLow{20}
\newlength\lunderset
\newlength\rulethick
\newcommand\att[4][1]{\setbox0=\hbox{#2}%
  \stackunder[#1\lunderset-\rulethick]{\strut#2}{\color{#3!#4}\rule{\wd0}{\rulethick}}}

\definecolor{att}{rgb}{0, 1, 0}
\definecolor{attb}{rgb}{1, 0, 0}

\makeatletter
  \providecommand*\setfloatlocations[2]{\@namedef{fps@#1}{#2}}
\makeatother
\setfloatlocations{figure}{htbp}
\setfloatlocations{table}{htbp}
\setfloatlocations{figure*}{htbp}
\setfloatlocations{table*}{htbp}

\newcommand\copyrighttext{%
  \footnotesize $\copyright$ 2020 IEEE.  Personal use of this material is permitted.  Permission from IEEE must be obtained for all other uses, in any current or future media, including reprinting/republishing this material for advertising or promotional purposes, creating new collective works, for resale or redistribution to servers or lists, or reuse of any copyrighted component of this work in other works.}
\newcommand\copyrightnotice{%
\begin{tikzpicture}[remember picture,overlay]
\node[anchor=south,yshift=0pt] at (current page.south)
{\parbox{\dimexpr\textwidth-\fboxsep-\fboxrule\relax}{\copyrighttext}};
\end{tikzpicture}%
}

\def\BibTeX{{\rm B\kern-.05em{\sc i\kern-.025em b}\kern-.08em
    T\kern-.1667em\lower.7ex\hbox{E}\kern-.125emX}}
\begin{document}
\title{Classification of cancer pathology reports: a large-scale comparative study}
\author{Stefano Martina, Leonardo Ventura, and Paolo Frasconi
\thanks{Paolo Frasconi was supported in part by the Italian Ministry of Education,  University, and Research under Grant 2017TWNMH2.}
\thanks{Stefano Martina is with University of Florence (email: stefano.martina@unifi.it).}
\thanks{Leonardo Ventura is with Institute for cancer research, prevention and clinical network (ISPRO), Florence (email: l.ventura@ispro.toscana.it).}
\thanks{Paolo Frasconi is with University of Florence (email: paolo.frasconi@unifi.it).}}

\maketitle
\copyrightnotice

\begin{abstract}
  We report about the application of state-of-the-art deep learning
  techniques to the automatic and interpretable assignment of ICD-O3
  topography and morphology codes to free-text cancer reports. We
  present results on a large dataset (more than 80\,000 labeled and
  1\,500\,000 unlabeled anonymized reports written in Italian and
  collected from hospitals in Tuscany over more than a decade) and
  with a large
  number of classes (134 morphological classes and 61 topographical
  classes). We compare alternative architectures in terms of
  prediction accuracy and interpretability and show that our best
  model achieves a multiclass accuracy of 90.3\% on topography site
  assignment and 84.8\% on morphology type assignment. We found that
  in this context hierarchical models are not better than flat models
  and that an element-wise maximum aggregator is slightly better than
  attentive models on site classification. Moreover, the maximum
  aggregator offers a way to interpret the classification process.
\end{abstract}


\section{Introduction}
\label{sec:introduction}
Cancer is a major concern worldwide, as it decreases the quality of
life and leads to premature mortality. In addition, it is one of the
most complex and difficult-to-treat
diseases, with significant social implications, both in terms of
mortality rate and in terms of costs associated with treatment and
disability~\cite{sullivan_delivering_2011,b._stewart_world_2014,desantis_cancer_2014,siegel_cancer_2016}.
Measuring the burden of disease is one of the main concerns of public
healthcare operators. Suitable measures are necessary to describe the general state
of population’s health, to establish public health goals and to
compare the national health status and performance of health systems
across countries. Furthermore, such studies are needed to assess the
allocation of health care and health research resources across disease
categories and to evaluate the potential costs and benefits of public
health interventions~\cite{brown_burden_2001}.

Cancer registries emerged during the last few decades as a strategic
tool to quantify the impact of the disease and to provide analytical
data to healthcare operators and decision makers.  Cancer registries
use administrative and clinical data sources in order to identify all
the new cancer diagnoses in a specific area and time period and
collect incidence records that provide details on the diagnosis and
the outcome of treatments.  Mining cancer registry datasets can help
towards the development of global surveillance
programs~\cite{tourassi_deep_2017} and can provide important insights
such as survivability~\cite{delen_predicting_2005}.  Although data
analysis software would best operate on structured representations of
the reports, pathologists normally enter data items as free text in
the local country language. This requires intelligent algorithms for
medical document information extraction, retrieval, and
classification, an area that has received significant attention in the
last few years (see, e.g.,~\cite{mujtaba_clinical_2019} for a recent
account and \cite{yim_natural_2016} for the specific case of cancer).

The study of intelligent algorithms is also motivated by the inherent
slowness of the cancer registration process, which is partially
based on manual revision, and which also requires the interpretation
of pathological reports written as free
text~\cite{jensen1991cancer5,bray2017cancer6,airtum_handbook_2008}. In 
practice, significant 
delays in data production and publication may occur. This weakens data
relevance for the purpose of assessing compliance with updated
regional recommended integrated case pathways, as well as for public
health purposes. Improving automated methods to generate a list of
putative incident cases and to automatically estimate process
indicators is thus an opportunity to perform an up-to-date evaluation
of cancer-care quality. In particular, machine learning techniques
like the ones presented in this paper could overcome the delay in
cancer case definition by the cancer registry and pave the way towards
powerful tools for obtaining indicators automatically and in a timely
fashion.

In our specific context, pathology reports can be classified according
to codes defined in the \ac{icdo3}
system~\cite{fritz_international_2000}, a specialization of the ICD
for the cancer domain which is internationally adopted as the standard
classification for topography and
morphology~\cite{airtum_handbook_2008}. The development of text
analysis tools specifically devoted to the automatic classification of
incidence records according to ICD-O3 codes has been addressed in a
number of previous papers (see Section~\ref{sec:related} below). Some
works have either focused on reasonably large datasets but using
simple linear classifiers based on bag-of-words representations of
text~\cite{jouhet_automated_2011,kavuluru_automatic_2013}. Most other
works have applied recent state-of-the-art deep learning
techniques~\cite{gao_hierarchical_2018,qiu_deep_2018} but using
smaller datasets and restricted to a partial set of tumors. A
remarkable exception is~\cite{alawad2020automatic} that applies
convolutional networks to a large dataset. Additionally, the use of
deep learning techniques usually requires accurate domain-specific
word vectors (embeddings of words in a vector space) that can be
derived from word co-occurrences in large corpora of unlabeled
text~\cite{mikolov_linguistic_2013,pennington_glove:_2014,devlin2018bert}. Large
medical corpora are easily available for English (e.g. PubMed) but not
necessarily for other languages.

To the best of our knowledge, the present work is the first to report
results on a large dataset ($> 80,000$ labeled reports for supervised
learning and $> 1,500,000$ unlabeled reports for pretraining word
vectors), with a large number of both topography and morphology
classes, and comparing several alternative state-of-the-art deep
learning techniques, namely \ac{gru}
\ac{rnn}~\cite{cho2014properties}, with and without
attention~\cite{bahdanau_neural_2014}, \ac{bert}~\cite{devlin2018bert}
and \ac{cnn}.  In particular, we are interested in evaluating on real
data the effectiveness of attention models, comparing them with a
simpler form based on max aggregation.  We also report an extensive
study on the interpretability of the trained classifiers.  Our results
confirm that recent deep learning techniques are very effective on
this task, with attentive \acp{gru} reaching a multiclass accuracy of
90.3\% on topography (61 classes) and 84.8\% on morphology (134
classes) but (1) hierarchical models does not achieve better accuracy
than using flat models, (2) the improvement over a simple support
vector machine classifier on bag-of-words is modest, (3) a simpler
aggregator of hidden representations taking element-wise maximum over
time improves slightly over (flat and hierarchical) attention models
for topography prediction while a flat attention model is better for
morphology task, and (4) the improvements of flat models over
hierarchical is stronger for difficult to learn rare classes.  We
additionally show that the element-wise maximum aggregator offers a
new alternative strategy for interpreting prediction results.

\section{Related Works}
\label{sec:related}
Early works for ICD-O3 code assignment were structured on rule-based
systems, where the code was assigned by creating a set of handcrafted
text search queries and combining results by standard Boolean
operators~\cite{crocetti_automatic_2004}. In order to prevent spurious
matches, rules need to be very specific, making it very difficult to
achieve a sufficiently high recall on future (unseen) cases.

Also more recent works employ rule-based approaches. Coden et
al. \cite{coden2009automatically} implemented a knowledge
representation model that they populate processing cancer pathology
reports with \ac{nlp} techniques. They performed categorization of
classes using rules based on syntactic structure. They also
experimented, without satisfactory results, machine learning
methods. They validate the model using a small corpus of 302 pathology
reports related to colon cancer obtaining an $F1$ score of $0.82$ for
primary tumor classification and $0.65$ for metastatic tumor. Nguyen et
al. \cite{nguyen2015assessing} developed a rule based system evaluated
on a set of $221$ pathology reports with $66$ full site classes (site
plus sub-site) and $94$
type classes. They obtained an $F1$ score of $0.61$ and $0.64$
respectively for site and type.

A number of studies reporting on the application of machine learning
to this problem have been published during the last decade. Direct
comparisons among these works are impossible due to the (not
surprising) lack of standard publicly available datasets and the
presence of 
heterogeneous details in the settings. Still, we highlight the main
differences among them in order to provide some 
background. In~\cite{jouhet_automated_2011}, the authors employed
support vector machine (SVM) and Naive Bayes classifiers on a small
dataset of $5\,121$ French pathology reports and a reduced number of
target classes (26 topographic classes and 18 morphological classes),
reporting an accuracy of 72.6\% on topography and 86.4\% on morphology
with SVM. A much larger dataset of $56\,426$ English reports from the
Kentucky Cancer Registry was later employed
in~\cite{kavuluru_automatic_2013}, where linear classifiers (SVM,
Naive Bayes, and logistic regression) were also compared but only on
the topography task and using 57, 42, and 14 classes (determined
considering classes that have at least respectively 50, 100, and 1000
examples). The authors 
reported a micro-averaged F1 measure of 90\% on 57 classes using SVM
with both
unigrams and bigrams. Still, the bag-of-words representations used by
these linear classifiers do not consider word order and are
unable to capture similarities and relations among words (which are
all represented by orthogonal vectors). Deep learning techniques are
known to overcome these limitations but were not applied to this
problem until very recently. In~\cite{qiu_deep_2018}, a \ac{cnn}
architecture fed by word vectors pretrained on 
PubMed was applied to a small corpus of 942 breast and lung cancer
reports in English with 12 topography classes; the authors demonstrate
the superiority of this approach compared to linear classifiers with
significant increases in both micro and macro F1
measures. In~\cite{gao_hierarchical_2018}, the same research group also
experimented on the same dataset using \acp{rnn}
with hierarchical attention~\cite{yang_hierarchical_2016}, obtaining
further improvements over the CNN architecture. Also the same research
group implemented in~\cite{alawad2020automatic} two CNN-based
multitask learning 
techniques and trained them on a big dataset of $95\,231$ pathology
reports 
($71\,223$ unique tumors) from the Louisiana Tumor Registry. The
models where trained on five tasks: topology main site (65 classes),
laterality (4 classes), behavior (3 classes), morphology type (63
classes), and morphology grade (5 classes). They reached a micro and
macro F1 score of respectively $0.944$ and 
$0.592$ for site prediction and respectively $0.811$ and $0.656$ for
type prediction.

Recent works investigated the interpretability of supervised machine
learning models. In \cite{ribeiro2016should}, a novel technique
called \emph{LIME} explains the prediction
of any classifier or regressor by locally approximating it.

\section{Materials and Methods}
\subsection{Dataset}
\label{sec:dataset}
We collected a set of $1\,592\,385$ anonymized anatomopathological exam results
from Tuscany region cancer registry in the period 1990-2014 for which we
obtained the approval from the institutional ethics committee\footnote{CEAV
14081\_oss 27/11/2018}. About $6\%$ 
of these records refer to a positive tumor
diagnosis and have topographical and morphological \ac{icdo3} labels,
determined by tumor registry experts. Other reports are associated
with non-cancerous tissues and with unlabeled records. When multiple
pathological records for the
same patient existed for the same tumor, cancer registry experts
selected the most 
informative report in order to assign the \ac{icdo3} code to that tumor
case, leaving a set of $94\,524$ labeled reports. In our dataset each
labeled report correspond to the primary report for a single tumor
case, thus the classification
was performed at report level.

The histological exam records consist of three free-text fields (not all
of them always filled-in) reporting tissue macroscopy, diagnosis,
and, in some cases, the patient's anamnesis. We found that field
semantics was not always used consistently and that the amount of
provided details varied significantly from extremely synthetic to very
detailed assessments of the morphology and the diagnosis. Field length
ranged from $0$ to $1\,368$ words, with lower, middle and upper
quartiles respectively 34, 62 and 134. For these reasons, we merged 
the three text fields
into a single text document. We did a case normalization converting
the letters to uppercase and we kept punctuation. We finally removed
duplicates (records that have the exact same text)
and reports labeled with extremely rare \ac{icdo3} codes (1048
samples that do not
appear in either training, validation, and test sets). In the end
we obtained 
a dataset suitable for supervised learning consisting of $85\,170$
labeled records ($\sim 99\%$ of them in the period 2004-2012).
We further split the records in sentences when using hierarchical models. For
this purpose we employed the \emph{spaCy} sentence segmentation
tool\footnote{\texttt{https://spacy.io/}}.

After preprocessing, our documents had on average length of 105 words
and contained on average 13 sentences (detailed distributions are
reported in~\cref{fig:wordsDist} of Appendix~\ref{app:distr}).  These
statistics indicate that reports tend to be much shorter than in other
studies. For example, in~\cite{alawad2020automatic} the average number
of words and sentences per document were 1290 and 117,
respectively\footnote{Note, however, than in that study about 76\% of
  the documents consisted of a single report, and the rest of
  concatenated reports (two reports in 17.7\% cases, three in 4.2\%
  cases, and four or more in 2.1\% cases).}. It is also the case that
language is often synthetic, rich in keywords, and poor in verbs (three
sample reports are shown in~\cref{tab:multiAttention1}).

\ac{icdo3} codes describe both topography (tumor site) and morphology. A
topographical \ac{icdo3} code is structured as \emph{Cmm.s} where
\emph{mm} represent the main site and \emph{s} the subsite. For
example, \emph{C50.2} is the code for the upper-inner quadrant
(\emph{2}) of breast (\emph{50}).  A morphological \ac{icdo3} code is
structured as \emph{tttt/b} where \emph{tttt} represent the cell type
and \emph{b} the tumor behavior (benign, uncertain, in-situ, malignant
primary site, malignant metastatic site). For example, \emph{8140/3}
is the code for an adenocarcinoma (\emph{adeno 8140}; \emph{carcinoma
  3}).  We defined two associated multi-class classification tasks (1)
main tumor site prediction (topography) and (2) type prediction
(morphology). The topography task only considers the first part of the
topographical \ac{icdo3} code, before the dot without the
sub-site. The morphology task only considers the
first part of the morphological \ac{icdo3} code, before the slash
without the behavior.
As shown in \cref{fig:classDist} (Appendix~\ref{app:distr}), our
dataset is highly unbalanced, with many of the 71 topographical and
435 morphological classes found in the data being very rare.
In an attempt to reduce bias in the estimated performance
(particularly for the macro F1 measure, see below), we removed classes
with less than five records in the test set, resulting in 61
topographical and 134 morphological classes. Even after these
removals, our tasks have no less classes than in previous works (the
most comprehensive previous study~\cite{alawad2020automatic} has 65
topographical classes and 63 histological classes).

In order to provide an evaluation that does not neglect possible
dataset shift issues over time (for example due to style changes or to
the evolving of oncology knowledge) we split train, validation, and
test data using a temporal criterion (based on record insertion date):
we used the most recent $20\%$ of data as the test
set ($17\,015$ records for site and
$16\,719$ for type, from March 2012 to March 2014), a similar amount of
the remaining most recent records 
as the validation set ($17\,007$ for site and
$16\,787$ for type, from December 2010 to March 2012), and the rest as the
training set ($50\,875$ for site and
$49\,436$ for type, before December 2010). Note that many other
previous studies have used a 
k-fold 
cross validation strategy, which is perhaps unavoidable when dealing
with small datasets.

\subsection{Plain models}
\label{sec:model}
In our setting, a dataset $\mathcal{D}=\{(\vect{x}^{(i)},y^{(i)})\}$
consists of variable length sequence vectors $\vect{x}^{(i)}$. For
$t=1,\dots,T^{(i)}$, $x^{(i)}_t$ is the $t$-th word in the $i$-th
document and $y^{(i)}\in\{1,\dots,K\}$ is the associated target
class. To simplify the notation in the subsequent text, the
superscripts are not used unless necessary. Sequences are denoted in
boldface. The \ac{gru}-based sequence classifiers\footnote{ In a set
  of preliminary experiments we found that \ac{lstm} did not improve
  over \ac{gru}.} used in
this work compute their predictions $f(\vect{x})$ as follows:
\begin{align}
  e_t &= E(x_t;\theta^e),\label{eq:embed}\\
  h^f_t &= F(e_t,h^f_{t-1};\theta^f),\label{eq:maxModelRL}\\  
  h^r_t &= R(e_t,h^r_{t+1};\theta^r),\label{eq:maxModelRR}\\
  u_t &= G(h_t;\theta^h),\label{eq:maxModelF}\\
  \phi &= A(\vect{u};\theta^a),\label{eq:aggregation}\\
  f(\vect{x}) &= g(\phi;\theta^c).\label{eq:maxModelC}
\end{align}
$E$ is an embedding function mapping words into $p$-dimensional real
vectors where embedding parameters $\theta^e$ can be either
pretrained and adjustable or fixed, see Section~\ref{sec:word-vectors}
below.  Functions $F$ and $R$ correspond to (forward and reverse)
dynamics that can be described in terms of several (possibly layered)
recurrent cells. Each vector $h_t=h^f_t\oplus h^r_t$
(the concatenation of $h^f_t$ and $h^r_t$) can be interpreted as latent
representations of the information contained at position $t$ in the
document. $G$ is an additional \ac{mlp} (with sigmoidal output units) mapping each latent vector into a
vector $u_t$ that can be seen as contextualized representation of the
word at position $t$. $A$ is an aggregation function that creates a
single $d$-dimensional representation vector for the entire sequence
and $g$ is a classification layer with softmax.
The parameters $\theta^f,\theta^r,\theta^h$, and $\theta^a$ (if
present) are determined by minimizing a loss function $\loss$
(categorical cross-entropy in our case) on training data.
Three possible choices for the aggregator function are described below.
\subsubsection{Concatenation}
  $\phi=(h^f_T,h^r_1)$. In this model, called \gru{} in the following, $G$ is the identity
  function and we simply take the extreme latent representations; in
  principle, these may be sufficient since they depend on the whole
  sequence due to bidirectional dynamics. However, note that this
  approach may require long-term dependencies to be effectively
  learned;
\subsubsection{Attention mechanism}
  $\phi = \sum_t a_t(\vect{u};\theta^a) u_t$.
In this model, called \softmax{} in the following, (scalar) attention
weight~\cite{bahdanau_neural_2014} are computed as
\begin{align*}
  c_t&=C(\vect{u};\theta^a),\\
  a_t(\vect{u};\theta^a) &= \frac{e^{\langle c, c_t\rangle}}
                           {\sum_{\tau=1}^T{e^{\langle c, c_\tau\rangle}}},\\
\end{align*}
where $C$ is a single layer that maps the representation $u_t$ of the
word to a hidden representation $c_t$. Then, the importance of the word is
measured as a similarity with a context vector $c$ that is learned
with the model and can be seen as an embedded representation of a
high level query as in memory networks~\cite{sukhbaatar2015end};
\subsubsection{Max pooling over time}
$\phi_j = \max_t u_{j,t}$. In this model~\cite{DBLP:journals/jmlr/CollobertWBKKK11,DBLP:conf/emnlp/Kim14}
(called \maxp{} in the following)
the sequence of representation vectors is treated as a bag and we apply a
form of multi-instance learning: each ``feature''
$\phi_j$ will be positive if at least one of $u_{j,1},\dots,u_{j,T}$
is positive (see also~\cite{tibo2017network}).  The resulting
classifier will find it easy to create decision rules predicting a
document as belonging to a certain class if a given set of
contextualized word representations are present and another given
set of contextualized word representations are absent in the
sequence. Note that this aggregator can also be interpreted as a
kind of hard attention mechanism where attention concentrates
completely on a single time step but the attended time step is
different for each feature $\phi_j$. As detailed in
Section~\ref{sec:interpretable}, a new model interpretation strategy
can be derived when using this aggregator.

\subsection{Interpretable model}\label{sec:interpretable}
An interpretable model can be used to assist the manual classification
process routinely performed in tumor registries and to explain the
proposed automatic classification for further human inspection. To
this end, the plain model (Eqs. \ref{eq:embed}--\ref{eq:maxModelC})
can be modified as follows:
\begin{align}
  e_t &= E(x_t;\theta^e),\\
  h^f_t &= F(e_t,h^f_{t-1};\theta^f),\\
  h^r_t &= R(e_t,h^r_{t+1};\theta^r),\\
  u_t &= G(h_t;\theta^h),\label{eq:maxiModelU}\\
  f(\vect{x}) &= A(\vect{u};\theta^a),
\end{align}
where $E$, $F$, $R$, $G$ and $A$ are defined as in \cref{sec:model}
and the size of $u_t$ is forced to equal the number of classes
so that each component $u_{j,t}$ of $u_t$
will be associated with the importance of words around position $t$
for class $j$.  This information can be used to interpret the model
decision. Preliminary experiments showed that the interpretation using
the attention aggregator was not satisfactory. Therefore in the
experiments we only report the interpretable model with the max
aggregator that we call \maxi{}. More details on the preliminary
experiments are reported in~\cite{martina2020}.  Besides accuracy, we
are also interested in the average agreement between \maxi{} and
\maxp{} (i.e. the fidelity of the interpretable classifier, see
Appendix~\ref{app:measures} for a definition).

\subsection{Hierarchical models}
\label{sec:modelh}
The last two models in \cref{sec:model} can be extended in a
hierarchical fashion,
as suggested in~\cite{yang_hierarchical_2016}. In this
case, data $\mathcal{D}=\{\vect{x}^{(i)},y^{(i)}\}$
consist of variable length sequences of sequence vectors
$\vect{x}^{(i)}$, where, for $s=1,\dots,S^{(i)}$ and
$t=1,\dots,T_s^{(i)}$, $x_{s,t}^{(i)}$ is the $t$-th word of the
$s$-th sentence in the 
$i$-th document, and $y^{(i)}\in\{1,\dots,K\}$ is the associated
target class. The prediction $f(\vect{x})$ is calculated as: 
\begin{align}
  e_{s,t} &= E(x_{s,t};\theta^e),\label{eq:embedH}\\
  h^f_{s,t} &= F(e_{s,t},h^f_{s,t-1};\theta^{f}),\label{eq:maxModelRLH}\\  
  h^r_{s,t} &= R(e_{s,t},h^r_{s,t+1};\theta^{r}),\label{eq:maxModelRRH}\\
  u_{s,t} &= G(h_{s,t};\theta^{h}),\label{eq:maxModelFH}\\
  \phi_s &= A(\vect{u}_s;\theta^{a}),\label{eq:aggregationH}\\
  \bar{h}^{f}_{s} &= \bar{F}(\phi_{s},\bar{h}^{f}_{s-1};\bar{\theta}^{f}),\label{eq:maxModelRLHS}\\  
  \bar{h}^{r}_{s} &= \bar{R}(\phi_{s},\bar{h}^{r}_{s+1};\bar{\theta}^{r}),\label{eq:maxModelRRHS}\\
  \bar{\phi} &= \bar{A}(\bar{\vect{h}};\bar{\theta}^{a}),\label{eq:aggregationHS}\\
  f(\vect{x}) &= g(\bar{\phi};\theta^c).\label{eq:maxModelCH}
\end{align}
As in the plain model, $E$ is an embedding function, $F$ and $R$
correspond to forward and reverse dynamics that process word
representations,
$h_{s,t}=h_{s,t}^f\oplus h_{s,t}^r$ is the latent representation of
the information contained at position $t$ of the $s$-th sentence,
$u_{s,t}$ is the contextualized representation of the word at position
$t$ of the $s$-th sentence, and $A$ is an aggregation function that
creates a single representation for the sentence. Furthermore, $\bar{F}$ and
$\bar{R}$ correspond to 
forward and reverse dynamics that process sentence representations,
and $\bar{A}$ is the aggregation function that creates a single
representation for the entire document. $\bar{h}_s=\bar{h}^f_s\oplus \bar{h}^r_s$
can be interpreted as the
latent representation of the information contained in the sentence $s$
for the document.
We call \maxh{} and \softmaxh{} the hierarchical versions of \maxp{}
and \softmax{}, respectively.

\subsection{Word Vectors}
\label{sec:word-vectors}
Most algorithms for obtaining word vectors are based on co-occurrences
in large text corpora. Co-occurrence can be measured either at the
word-document level (e.g.\ using latent semantic analysis) or at the
word-word level (e.g.\ using word2vec~\cite{mikolov_linguistic_2013}
or \ac{glove}~\cite{pennington_glove:_2014}). It is a common practice to
use pre-compiled libraries of word vectors trained on
several billion tokens extracted from various sources such as
Wikipedia, the English Gigaword 5, Common Crawl, or Twitter. These
libraries are generally conceived for general purpose applications and
are only available for the English language. Reports in cancer
registries, however, are normally written in the local language and
make extensive usage of a very specific domain terminology. In fact
they can be considered \emph{sublanguages} with a specific vocabulary
usage
and with peculiar sentence construction rules that differ from the
normal construction rules \cite{spyns1996natural}.

Another approach is to employ a \ac{lm} that models language as a
sequence of characters instead of words. In particular, in the
\emph{Flair} framework~\cite{akbik2018contextual}, the internal states
of a trained character level \ac{lm} are used to produce
\emph{contextual string} word embeddings.

\subsection{Baselines}
\label{sec-baselines}

\subsubsection{Linear classifiers}
The classic approach is to employ bag-of-words
representations of textual documents.
Vector representations of documents are easily
derived from bags-of-words either by using indicator vectors or taking
into account the number of occurrences of each word using the
\ac{tfidf}\cite{manning_introduction_2008}. In those representations,
frequent and non-specific terms receive a lower weight.

Bag-of-words representations (including those employing bigrams or
trigrams) enable the application of linear text classifiers, such as
\ac{nb}, \ac{svm} \cite{cortes-support-1995}, or boosted tree
classifiers~\cite{chen2016xgboost}. Those representations suffer two
fundamental 
problems: first, the relative order of terms in the documents is lost,
making it impossible to take advantage of the syntactic structure of
the sentences; second, distinct words have an orthogonal
representation even when they are semantically close. Word vectors can
be used to address the second 
limitation and also allow us to take advantage of unlabeled data,
which can be typically be obtained in large amounts and with little
cost.

\subsubsection{\ac{cnn}}
\acf{cnn} can be successfully employed in the context of sentence
classification~\cite{DBLP:conf/emnlp/Kim14}.
The \ac{cnn} model that we trained in
our work is a slight variant of the architecture
in~\cite{DBLP:conf/emnlp/Kim14}. The original architecture produces
features maps applying convolutional filters on the sequence of word
vectors followed by a max pooling and the classification. We used
three convolutional layers with filter size of 3, 4 and 5. Moreover we
added a linear layer between the word vectors and the convolutional
layers. We fine-tuned hyperparameters for the output size of the
linear layer and the number of convolutional filters. The input size
of the linear layer is the same as the word vector size.

\subsubsection{\acs{bert}}
\ac{bert} \cite{devlin2018bert} is a recent model that represents the
state of the art in many \ac{nlp} related tasks
\cite{chatterjee2019semeval,hu2019introductory,lee2019biobert,tshitoyan2019unsupervised}.
It is a
bi-directional pre-training model backboned by the Transformer Encoder
\cite{vaswani2017attention}. It is an attention-based technique that
learns context-dependent word representation on large unlabeled
corpora, and then the model is fine tuned end-to-end on specific labeled
tasks. During pre-training, the model is trained
on unlabeled data over two different tasks. In \ac{mlm} some tokens
are masked and the model is trained to predict those token based on
the context. In \ac{nsp} the model is trained to understand the
relationship between sentences predicting if two sentences are actually
consecutive or if they where randomly replaced (with 50\%
probability).

In our work we pre-trained \ac{bert} using the same set of 1.5
million 
unlabeled records that we used to train word embeddings (see
\cref{sec:experiments} for details). Then we fine tuned \ac{bert} with the
specific topography and morphology prediction tasks.

\subsection{Hyperparameters}\label{sec:hyperparameters}
All deep models (\gru{}, \maxp{}, \softmax{}, \maxi{}, \maxh{}, and
\softmaxh{}) were trained by minimizing the categorical cross entropy
loss with Adam \cite{kingma2014adam} with an initial learning rate of
$0.001$ and minibatches of $32$ samples.  The remaining
hyperparameters (including $C$ for SVM) were obtained by grid search using the validation
accuracy as the objective (see Appendix~\ref{app:hyper} for optimal values and details on the hyperparameter space). In
particular, we tuned hyperparameters in
\eqref{eq:embed}~-~\eqref{eq:maxModelC} and
\eqref{eq:embedH}~-~\eqref{eq:maxModelCH}, which control the structure
of the model.

$\xi^e$ is associated with the embedding layer $E$ and in our case
refers to \ac{glove} 
hyperparameters~\cite{pennington_glove:_2014}. With an intrinsic
evaluation, we found that the better configuration was $60$ for the vector
size, $15$ for the window size, and $50$ iterations. We constructed
sets of couples of related words, i.e. 11, 12, 11, 7 and 92 couples
for respectively the benign-malignant, benign-tissue,
malignant-tissue, morphology-site and singular-plural
relations. For example,
\emph{fibroma}, \emph{fibrosarcoma} and \emph{lipoma},
\emph{liposarcoma} for the benign-malignant relation and \emph{fibroma},
\emph{connective} and \emph{lipoma} \emph{adipose} for the cancer-tissue
relation. We then used those sets to evaluate if the semantic relations are
captured by 
linear substructures in the space of the embeddings, e.g. we measure
if $E(\emph{fibrosarcoma})-E(\emph{fibroma})+E(\emph{lipoma})\approx
E(liposarcoma)$ for the benign-malignant relation and
$E(\emph{fibroma})-E(\emph{connective})+E(\emph{adipose})\approx 
E(lipoma)$ for the cancer-tissue relation.
We confirmed the
parameters with an extrinsic evaluation on the best model by grid
search in the space of $[2,\dots,20]$ for window size and
$[40,\dots,300]$ for vector dimension.

$\xi^f$,
$\xi^r$, $\bar{\xi}^{f}$, and $\bar{\xi}^{r}$ define the number of
\ac{gru} layers ($\xi_{(l)}$) and the number of unit per each layer
($\xi_{(d)}$) respectively for $F$, $R$, $\bar{F}$, and $\bar{R}$. 
$G$ is a \ac{mlp}, $\xi^h$ controls the number of
layers ($\xi^h_{(l)}$) and their size ($\xi^h_{(d)}$). Regarding $F$,
$R$, and $G$, we decided to 
have all the stacked layer with 
the same size to limit the
hyperparameters space. $\xi^a$ and $\bar{\xi}^a$ control the
kind of aggregating function of $A$ and $\bar{A}$ respectively and, in case
of \emph{attention}, 
it controls the size of the attention layer ($\xi^a_{(d)}$). Finally,
$\xi^c$ controls the 
data-dependent output size of $g$.

\section{Results}\label{sec:experiments}
In the experiments reported below word vectors were computed by
\ac{glove}~\cite{pennington_glove:_2014} trained on our set of 1.5 millions
unlabeled records.
In a set of preliminary experiments, we also compared the best model that
we obtained using \ac{glove} embeddings against the same model trained
using Flair embeddings obtained using a \ac{lm} trained on the same
unlabeled records. Although Flair has the potential advantage of
robustness with respect to typos and spelling variants, extrinsic
results on the topography and the morphology tasks did not show any
advantages over \ac{glove}. For example test-set accuracy attained on topography
by the best model, \maxp{}, were slightly worse with Flair embeddings
(89.9\%) than with \ac{glove} embeddings (90.3\%) (the latter is reported
in Table~\ref{tab:results_site}).

\begin{table}
  \centering
  \caption{Topography site prediction (61 classes), significance
    against \maxp{} ($^*$: $p<10^{-2}$; $^{**}$: $p<10^{-3}$; $^{***}$: $p<10^{-4}$)}
  \label{tab:results_site}
  \begin{tabular}{ccccc}
    \hline
                  & Accuracy                       & Top 3 Acc.        & Top 5 Acc.        & MacroF1                                               \\
    \hline
    \svm{}        & 89.7$^{**}$\a                  & 95.9$^{***}$      & 96.8$^{***}$      & 60.0\aaa                                              \\
    \hline
    \cnn{}        & 89.2$^{***}$                   & 96.0$^{***}$      & 97.6$^{***}$      & 55.3$^{***}$                                          \\
    \gru{}        & 89.9$^{*}$\aa                  & 96.5\aaa          & 97.7$^{***}$      & 58.3$^{**}$\a                                         \\
    \bert{}       & 89.9$^{*}$\aa                  & 96.3$^{*}$\aa     & 97.8$^{*}$\aa     & 56.6$^{*}$\aa                                         \\
    \hline
    \maxi{}       & 88.0$^{***}$                   & 95.4$^{***}$      & 96.2$^{***}$      & 46.1$^{***}$                                          \\
    \maxh{}       & 89.9$^{*}$\aa                  & 96.2$^{***}$      & 97.8$^{*}$\aa     & 58.8$^{*}$\aa                                         \\
    \softmaxh{}   & 89.9\aaa                       & 96.3$^{**}$\a     & 97.7$^{**}$\a     & 58.0$^{**}$\a                                         \\
    \maxp{}       & \textbf{90.3}\aaa              & \textbf{96.6}\aaa & \textbf{98.1}\aaa & \textbf{61.9}\aaa                                     \\
    \softmax{}    & 90.1\aaa                       & 96.2$^{***}$      & 97.6$^{***}$      & 60.0\aaa                                              \\
    \hline
  \end{tabular}
\end{table}
\begin{table}
  \centering
  \caption{Morphology type prediction (134 classes), significance
    against \maxp{} ($^*$: $p<10^{-2}$; $^{**}$: $p<10^{-3}$; $^{***}$: $p<10^{-4}$)}
  \label{tab:results_morpho}
  \begin{tabular}{ccccc}
    \hline
                  & Accuracy                       & Top 3 Acc.        & Top 5 Acc.        & Macro F1                                              \\
    \hline
    \svm{}        & 82.4$^{***}$                   & 94.0$^{***}$      & 95.6$^{***}$      & 53.7$^{**}$\a                                         \\
    \hline
    \cnn{}        & 83.3$^{***}$                   & 94.4$^{***}$      & 96.7\aaa          & 55.0$^{**}$\a                                         \\
    \gru{}        & 83.3$^{***}$                   & 94.6$^{*}$\aa     & 96.6$^{*}$\aa     & 55.2$^{**}$\a                                         \\
    \bert{}       & 84.3\aaa                       & 93.2$^{***}$      & 94.9$^{***}$      & 51.1$^{***}$                                          \\    
    \hline
    \maxi{}       & 73.4$^{***}$                   & 91.0$^{***}$      & 93.6$^{***}$      & 31.3$^{***}$                                          \\
    \maxh{}       & 83.7$^{***}$                   & 94.4$^{***}$      & 96.4$^{***}$      & 54.5$^{*}$\aa                                         \\
    \softmaxh{}   & 83.7$^{***}$                   & 94.4$^{***}$      & 96.2$^{***}$      & 57.5\aaa                                              \\
    \maxp{}       & 84.6\aaa                       & \textbf{95.0}\aaa & \textbf{96.9}\aaa & 59.2\aaa                                              \\
    \softmax{}    & \textbf{84.8}\aaa              & 94.9\aaa          & \textbf{96.9}\aaa & \textbf{61.3}\aaa                                     \\
    \hline
  \end{tabular}
\end{table}
\begin{table}
  \centering
  \caption{Macro F1 measure by groups of class frequency, signific.
    against \maxp{} ($^*$: $p<10^{-2}$; $^{**}$: $p<10^{-3}$; $^{***}$: $p<10^{-4}$)}
  \label{tab:results_difficulty}
  \begin{tabular}{cccc|ccc}
    \hline
                  & \multicolumn{3}{c}{Topography} & \multicolumn{3}{c}{Morphology}                                                                \\
                  & easy                           & avg.              & hard              & easy            & avg.             & hard             \\
                  & (4 cls)                        & (18 cls)          & (39 cls)          & (5 cls)         & (18 cls)         & (111 cls)        \\
    \hline
    \svm{}        & 95.7$^{*}$                     & \textbf{86.9}\aa  & 50.9\aaa          & 90.5\a          & 68.6\aa          & 48.4$^{*}$\aa          \\
    \hline
    \cnn{}        & 95.6\a                         & 71.0$^{**}$       & 43.1$^{***}$      & 91.7$^{*}$      & 70.5\aa          & 49.2$^{**}$\a     \\
    \gru{}        & \textbf{96.1}\a                & 72.2\aa           & 48.0$^{*}$\aa     & 91.4\a          & 71.6\aa          & 49.7$^{**}$\a     \\
    \bert{}       & 95.7\a                         & 73.2\aa           & 44.9$^{*}$\aa     & \textbf{92.9}\a & \textbf{74.4}\aa & 43.9$^{***}$      \\    
    \hline
    \maxi{}       & 95.0\a                         & 66.6\aa           & 31.4$^{***}$      & 87.1\a          & 41.9$^{**}$      & 25.1$^{***}$      \\
    \maxh{}       & 95.8\a                         & 72.4\aa           & 48.8$^{*}$\aa     & 92.7\a          & 71.8\aa          & 48.8$^{*}$\aa          \\
    \softmaxh{}   & 96.0\a                         & 73.1\aa           & 47.1$^{**}$\a     & 91.9\a          & 72.3\aa          & 52.6\aaa          \\
    \maxp{}       & 96.0\a                         & 73.3\aa           & \textbf{53.1}\aaa & 92.7\a          & 72.3\aa          & 53.8\aaa          \\
    \softmax{}    & 96.0\a                         & 73.1\aa           & 50.3\aaa          & 92.8\a          & 72.3\aa          & \textbf{56.7}\aaa \\
    \hline
  \end{tabular}
\end{table}
In~\cref{tab:results_site} and \cref{tab:results_morpho} we summarize the
results of different models on test data in terms of multiclass accuracy (or,
equivalently, micro-averaged F1 measure), top-$\ell$ accuracy (if the correct
class appears within the top $\ell$ predictions) for $\ell=3$ and $\ell=5$,
and macro-averaged F1 measure (see Appendix~\ref{app:measures} for definitions).
Significance (each method against \maxp{}) is reported with asterisks
in the tables and was assessed with a one-sided McNemar
test~\cite{Dietterich_1998} for accuracy and with a one-sided
macro T-test~\cite{yang1999re} for F1 score.

Collecting results for all the models (for a single hyperparameters
configuration and excluding the training of word vectors and \bert{})
required approximately 11 hours on a GeForce RTX 2080 Ti
GPU\footnote{The source code for the experiments is available at the following address:\\\url{https://github.com/trianam/cancerReportsClassification}}.
In~\cref{tab:results_difficulty} we
report F1 score averaged on different subsets of classes. We
consider a class \emph{easy} if it has more than $1000$ examples in
the test set, \emph{average} if it has between $100$ and $1000$
examples, and \emph{hard} if it has less than $100$ examples.

\begin{figure*}
  \centering
  \ttfamily
  \scriptsize
  \begin{tabular}{|c|c|c|c|}
    \hline
    { \sffamily Class}&{ \sffamily Relevant classes}& {\sffamily Document text with highlighted words}
    &{ \sffamily English translation (by the authors)}\\
    \hline
    61&\begin{minipage}{\attTableIcdoWidth}\att{\att{\att{61}{red}{\attHigh}}{red}{\attMid}}{red}{\attLow} (PROSTATE \\\\[-1.2em] \rule{2em}{0pt}GLAND)\end{minipage}&\begin{minipage}{\attTableTextWidth}\att{DISOMOGENICITA}{red}{0} \att{'}{red}{0} \att{DIFFUSE}{red}{0} \att{.}{red}{0} \att{PSA}{red}{\attHigh} \att{NON}{red}{0} \att{PERVENUTO}{red}{0} \att{.}{red}{0} \att{ADENOCARCINOMA}{red}{0} \att{PROSTATICO}{red}{\attHigh} \att{A}{red}{0} \att{GRADO}{red}{0} \att{DI}{red}{0} \att{DIFFERENZIAZIONE}{red}{0} \att{MEDIO}{red}{0} \att{-}{red}{0} \att{BASSO}{red}{0} \att{(}{red}{0} \att{GLEASON}{red}{\attHigh} \att{3}{red}{0} \att{+}{red}{0} \att{4}{red}{0} \att{)}{red}{0} \att{NEI}{red}{0} \att{PRELIEVI}{red}{0} \att{DI}{red}{0} \att{CUI}{red}{0} \att{AI}{red}{0} \att{NN}{red}{0} \att{.}{red}{0} \att{2}{red}{0} \att{E}{red}{0} \att{3}{red}{0} \att{.}{red}{0} \att{AGOBIOPSIA}{red}{0} \att{DELLA}{red}{0} \att{PROSTATA}{red}{\attHigh} \att{:}{red}{0} \att{1}{red}{0} \att{)}{red}{0} \att{1}{red}{0} \att{PRELIEVO}{red}{0} \att{LL}{red}{0} \att{DX}{red}{0} \att{.}{red}{0} \att{2}{red}{0} \att{)}{red}{0} \att{2}{red}{0} \att{PRELIEVI}{red}{0} \att{ML}{red}{0} \att{DX}{red}{0} \att{.}{red}{0} \att{3}{red}{0} \att{)}{red}{0} \att{2}{red}{0} \att{PRELIEVI}{red}{0} \att{M}{red}{0} \att{DX}{red}{0} \att{.}{red}{0} \att{4}{red}{0} \att{)}{red}{0} \att{1}{red}{0} \att{PRELIEVO}{red}{0} \att{M}{red}{0} \att{SX}{red}{0} \att{.}{red}{0} \att{5}{red}{0} \att{)}{red}{0} \att{2}{red}{0} \att{PRELIEVI}{red}{0} \att{ML}{red}{0} \att{SX}{red}{0} \att{.}{red}{0} \att{6}{red}{0} \att{)}{red}{0} \att{1}{red}{0} \att{PRELIEVO}{red}{0} \att{LL}{red}{0} \att{SX}{red}{0} \att{.}{red}{0} \att{7}{red}{0} \att{)}{red}{0} \att{1}{red}{0} \att{PRELIEVO}{red}{0} \att{TRANSIZIONALE}{red}{\attMid} \att{SX}{red}{0} \att{.}{red}{0} \att{8}{red}{0} \att{)}{red}{0} \att{1}{red}{0} \att{PRELIEVO}{red}{0} \att{TRANSIZIONALE}{red}{\attLow} \att{DX}{red}{0} \att{.}{red}{0}\end{minipage}&\begin{minipage}{\attTableTextWidth}\att{DIFFUSE}{red}{0} \att{DISHOMOGENEITY}{red}{0} \att{.}{red}{0} \att{PSA}{red}{\attHigh} \att{NOT}{red}{0} \att{RECEIVED}{red}{0} \att{.}{red}{0} \att{PROSTATIC}{red}{\attHigh} \att{ADENOCARCINOMA}{red}{0} \att{OF}{red}{0} \att{INTERMEDIATE}{red}{0} \att{-}{red}{0} \att{LOW}{red}{0} \att{GRADE}{red}{0} \att{OF}{red}{0} \att{DIFFERENTIATION}{red}{0} \att{(}{red}{0} \att{GLEASON}{red}{\attHigh} \att{3}{red}{0} \att{+}{red}{0} \att{4}{red}{0} \att{)}{red}{0} \att{IN}{red}{0} \att{SAMPLES}{red}{0} \att{AT}{red}{0} \att{N}{red}{0} \att{.}{red}{0} \att{2}{red}{0} \att{AND}{red}{0} \att{3}{red}{0} \att{.}{red}{0} \att{NEEDLE}{red}{0} \att{BIOPSY}{red}{0} \att{OF}{red}{0} \att{THE}{red}{0} \att{PROSTATE}{red}{\attHigh} \att{:}{red}{0} \att{1}{red}{0} \att{)}{red}{0} \att{1}{red}{0} \att{RIGHT}{red}{0} \att{LL}{red}{0} \att{SAMPLE}{red}{0} \att{.}{red}{0} \att{2}{red}{0} \att{)}{red}{0} \att{2}{red}{0} \att{RIGHT}{red}{0} \att{ML}{red}{0} \att{SAMPLES}{red}{0} \att{.}{red}{0} \att{3}{red}{0} \att{)}{red}{0} \att{2}{red}{0} \att{RIGHT}{red}{0} \att{M}{red}{0} \att{SAMPLES}{red}{0} \att{.}{red}{0} \att{4}{red}{0} \att{)}{red}{0} \att{1}{red}{0} \att{LEFT}{red}{0} \att{M}{red}{0} \att{SAMPLE}{red}{0} \att{.}{red}{0} \att{5}{red}{0} \att{)}{red}{0} \att{2}{red}{0} \att{LEFT}{red}{0} \att{ML}{red}{0} \att{SAMPLES}{red}{0} \att{.}{red}{0} \att{6}{red}{0} \att{)}{red}{0} \att{1}{red}{0} \att{LEFT}{red}{0} \att{LL}{red}{0} \att{SAMPLE}{red}{0} \att{.}{red}{0} \att{7}{red}{0} \att{)}{red}{0} \att{1}{red}{0} \att{LEFT}{red}{0} \att{TRANSITIONAL}{red}{\attMid} \att{SAMPLE}{red}{0} \att{.}{red}{0} \att{8}{red}{0} \att{)}{red}{0} \att{1}{red}{0} \att{RIGHT}{red}{0} \att{TRANSITIONAL}{red}{\attLow} \att{SAMPLE}{red}{0} \att{.}{red}{0}\end{minipage} \\
    \hline
    20&\begin{minipage}{\attTableIcdoWidth}\att{\att{\att{18}{red}{\attHigh}}{red}{\attMid}}{red}{\attLow} (COLON) \\\\\att{\att{\att{20}{green}{\attHigh}}{green}{\attMid}}{green}{\attLow} (RECTUM) \\\\\att{\att{\att{21}{blue}{\attHigh}}{blue}{\attMid}}{blue}{\attLow} (ANUS AND \\\\[-1.2em] \rule{1.5em}{0pt}ANAL CANAL)\end{minipage}&\begin{minipage}{\attTableTextWidth}\att{\att{\att{ISOLATI}{red}{0}}{green}{0}}{blue}{0} \att{\att{\att{FRAMMENTI}{red}{0}}{green}{0}}{blue}{0} \att{\att{\att{RIFERIBILI}{red}{0}}{green}{0}}{blue}{0} \att{\att{\att{AD}{red}{0}}{green}{0}}{blue}{0} \att{\att{\att{ADENOMA}{red}{0}}{green}{0}}{blue}{0} \att{\att{\att{TUBULARE}{red}{\attHigh}}{green}{0}}{blue}{0} \att{\att{\att{INTESTINALE}{red}{\attHigh}}{green}{0}}{blue}{0} \att{\att{\att{DI}{red}{0}}{green}{0}}{blue}{0} \att{\att{\att{ALTO}{red}{0}}{green}{0}}{blue}{0} \att{\att{\att{GRADO}{red}{0}}{green}{0}}{blue}{0} \att{\att{\att{.}{red}{0}}{green}{0}}{blue}{0} \att{\att{\att{FRAMMENTI}{red}{0}}{green}{0}}{blue}{0} \att{\att{\att{(}{red}{0}}{green}{0}}{blue}{0} \att{\att{\att{NR}{red}{0}}{green}{0}}{blue}{0} \att{\att{\att{.}{red}{0}}{green}{0}}{blue}{0} \att{\att{\att{2}{red}{0}}{green}{0}}{blue}{0} \att{\att{\att{)}{red}{0}}{green}{0}}{blue}{0} \att{\att{\att{DI}{red}{0}}{green}{0}}{blue}{0} \att{\att{\att{POLIPO}{red}{\attHigh}}{green}{\attMid}}{blue}{0} \att{\att{\att{PEDUNCOLATO}{red}{\attHigh}}{green}{0}}{blue}{0} \att{\att{\att{A}{red}{0}}{green}{0}}{blue}{0} \att{\att{\att{20}{red}{0}}{green}{0}}{blue}{0} \att{\att{\att{CM}{red}{0}}{green}{0}}{blue}{0} \att{\att{\att{DALL}{red}{\attMid}}{green}{0}}{blue}{0} \att{\att{\att{'}{red}{0}}{green}{0}}{blue}{0} \att{\att{\att{ORIFIZIO}{red}{0}}{green}{\attHigh}}{blue}{\attHigh} \att{\att{\att{ANALE}{red}{0}}{green}{\attHigh}}{blue}{\attHigh} \att{\att{\att{.}{red}{0}}{green}{\attLow}}{blue}{0} \att{\att{\att{(}{red}{0}}{green}{0}}{blue}{0} \att{\att{\att{ESEGUITA}{red}{0}}{green}{0}}{blue}{0} \att{\att{\att{COLORAZIONE}{red}{0}}{green}{0}}{blue}{0} \att{\att{\att{EMATOSSILINA}{red}{0}}{green}{0}}{blue}{0} \att{\att{\att{-}{red}{0}}{green}{0}}{blue}{0} \att{\att{\att{EOSINA}{red}{0}}{green}{0}}{blue}{0} \att{\att{\att{)}{red}{0}}{green}{0}}{blue}{0} \att{\att{\att{.}{red}{0}}{green}{0}}{blue}{0}\end{minipage}&\begin{minipage}{\attTableTextWidth}\att{\att{\att{ISOLATED}{red}{0}}{green}{0}}{blue}{0} \att{\att{\att{FRAGMENTS}{red}{0}}{green}{0}}{blue}{0} \att{\att{\att{ATTRIBUTABLES}{red}{0}}{green}{0}}{blue}{0} \att{\att{\att{TO}{red}{0}}{green}{0}}{blue}{0} \att{\att{\att{HIGH}{red}{0}}{green}{0}}{blue}{0} \att{\att{\att{DEGREE}{red}{0}}{green}{0}}{blue}{0} \att{\att{\att{INTESTINAL}{red}{\attHigh}}{green}{0}}{blue}{0} \att{\att{\att{TUBULAR}{red}{\attHigh}}{green}{0}}{blue}{0} \att{\att{\att{ADENOMA}{red}{0}}{green}{0}}{blue}{0} \att{\att{\att{.}{red}{0}}{green}{0}}{blue}{0} \att{\att{\att{FRAGMENTS}{red}{0}}{green}{0}}{blue}{0} \att{\att{\att{(}{red}{0}}{green}{0}}{blue}{0} \att{\att{\att{NR}{red}{0}}{green}{0}}{blue}{0} \att{\att{\att{.}{red}{0}}{green}{0}}{blue}{0} \att{\att{\att{2}{red}{0}}{green}{0}}{blue}{0} \att{\att{\att{)}{red}{0}}{green}{0}}{blue}{0} \att{\att{\att{OF}{red}{0}}{green}{0}}{blue}{0} \att{\att{\att{PEDUNCULATED}{red}{\attHigh}}{green}{0}}{blue}{0} \att{\att{\att{POLYPUS}{red}{\attHigh}}{green}{\attMid}}{blue}{0} \att{\att{\att{AT}{red}{0}}{green}{0}}{blue}{0} \att{\att{\att{20}{red}{0}}{green}{0}}{blue}{0} \att{\att{\att{CM}{red}{0}}{green}{0}}{blue}{0} \att{\att{\att{FROM}{red}{0}}{green}{0}}{blue}{0} \att{\att{\att{THE}{red}{\attMid}}{green}{0}}{blue}{0} \att{\att{\att{ANAL}{red}{0}}{green}{\attHigh}}{blue}{\attHigh} \att{\att{\att{ORIFICE}{red}{0}}{green}{\attHigh}}{blue}{\attHigh} \att{\att{\att{.}{red}{0}}{green}{\attLow}}{blue}{0} \att{\att{\att{(}{red}{0}}{green}{0}}{blue}{0} \att{\att{\att{PERFORMED}{red}{0}}{green}{0}}{blue}{0} \att{\att{\att{HEMATOXYLIN}{red}{0}}{green}{0}}{blue}{0} \att{\att{\att{-}{red}{0}}{green}{0}}{blue}{0} \att{\att{\att{EOSIN}{red}{0}}{green}{0}}{blue}{0} \att{\att{\att{COLORING}{red}{0}}{green}{0}}{blue}{0} \att{\att{\att{)}{red}{0}}{green}{0}}{blue}{0} \att{\att{\att{.}{red}{0}}{green}{0}}{blue}{0}\end{minipage} \\
    \hline
    34&\begin{minipage}{\attTableIcdoWidth}\att{\att{\att{34}{red}{\attHigh}}{red}{\attMid}}{red}{\attLow} (BRONCHUS \\\\[-1.2em] \rule{2em}{0pt}AND LUNG) \\\\\att{\att{\att{56}{green}{\attHigh}}{green}{\attMid}}{green}{\attLow} (OVARY) \\\\\att{\att{\att{67}{blue}{\attHigh}}{blue}{\attMid}}{blue}{\attLow} (BLADDER) \\\\\att{\att{\att{80}{black}{\attHigh}}{black}{\attMid}}{black}{\attLow} (UNKNOWN \\\\[-1.2em] \rule{2em}{0pt}PRIMARY \\\\[-1.2em] \rule{2em}{0pt}SITE)\end{minipage}&\begin{minipage}{\attTableTextWidth}\att{\att{\att{\att{VERSAMENTO}{red}{0}}{green}{0}}{blue}{0}}{black}{0} \att{\att{\att{\att{PLEURICO}{red}{\attHigh}}{green}{\attMid}}{blue}{0}}{black}{0} \att{\att{\att{\att{SX}{red}{\attLow}}{green}{0}}{blue}{0}}{black}{0} \att{\att{\att{\att{DI}{red}{0}}{green}{0}}{blue}{0}}{black}{0} \att{\att{\att{\att{N}{red}{0}}{green}{0}}{blue}{0}}{black}{0} \att{\att{\att{\att{.}{red}{0}}{green}{0}}{blue}{0}}{black}{0} \att{\att{\att{\att{D}{red}{0}}{green}{0}}{blue}{0}}{black}{0} \att{\att{\att{\att{.}{red}{0}}{green}{0}}{blue}{0}}{black}{0} \att{\att{\att{\att{D}{red}{0}}{green}{0}}{blue}{0}}{black}{0} \att{\att{\att{\att{.}{red}{0}}{green}{0}}{blue}{0}}{black}{0} \att{\att{\att{\att{E}{red}{0}}{green}{0}}{blue}{0}}{black}{0} \att{\att{\att{\att{ADDENSAMENTI}{red}{\attLow}}{green}{0}}{blue}{0}}{black}{0} \att{\att{\att{\att{POLMONARI}{red}{\attHigh}}{green}{0}}{blue}{0}}{black}{0} \att{\att{\att{\att{DI}{red}{0}}{green}{0}}{blue}{0}}{black}{0} \att{\att{\att{\att{N}{red}{0}}{green}{0}}{blue}{0}}{black}{0} \att{\att{\att{\att{.}{red}{0}}{green}{0}}{blue}{0}}{black}{0} \att{\att{\att{\att{D}{red}{0}}{green}{0}}{blue}{0}}{black}{0} \att{\att{\att{\att{.}{red}{0}}{green}{0}}{blue}{0}}{black}{0} \att{\att{\att{\att{D}{red}{0}}{green}{0}}{blue}{0}}{black}{0} \att{\att{\att{\att{.}{red}{0}}{green}{0}}{blue}{0}}{black}{0} \att{\att{\att{\att{,}{red}{0}}{green}{0}}{blue}{0}}{black}{0} \att{\att{\att{\att{NODULI}{red}{0}}{green}{0}}{blue}{0}}{black}{0} \att{\att{\att{\att{PARETE}{red}{0}}{green}{0}}{blue}{\attLow}}{black}{0} \att{\att{\att{\att{ADDOMINALE}{red}{0}}{green}{0}}{blue}{0}}{black}{0} \att{\att{\att{\att{.}{red}{0}}{green}{0}}{blue}{0}}{black}{0} \att{\att{\att{\att{INFILTRAZIONE}{red}{0}}{green}{0}}{blue}{0}}{black}{0} \att{\att{\att{\att{CANCERIGNA}{red}{\attHigh}}{green}{0}}{blue}{0}}{black}{\attLow} \att{\att{\att{\att{DEGLI}{red}{\attHigh}}{green}{0}}{blue}{0}}{black}{\attHigh} \att{\att{\att{\att{STROMI}{red}{0}}{green}{0}}{blue}{0}}{black}{\attHigh} \att{\att{\att{\att{CONNETTIVO}{red}{0}}{green}{0}}{blue}{0}}{black}{0} \att{\att{\att{\att{-}{red}{0}}{green}{0}}{blue}{0}}{black}{0} \att{\att{\att{\att{ADIPOSI}{red}{0}}{green}{0}}{blue}{0}}{black}{\attMid} \att{\att{\att{\att{.}{red}{\attLow}}{green}{0}}{blue}{0}}{black}{0} \att{\att{\att{\att{IMMUNOISTOCHIMICA}{red}{\attHigh}}{green}{0}}{blue}{0}}{black}{\attHigh} \att{\att{\att{\att{:}{red}{\attLow}}{green}{0}}{blue}{0}}{black}{\attHigh} \att{\att{\att{\att{CK7}{red}{\attHigh}}{green}{0}}{blue}{0}}{black}{\attHigh} \att{\att{\att{\att{+}{red}{\attHigh}}{green}{0}}{blue}{0}}{black}{0} \att{\att{\att{\att{,}{red}{\attMid}}{green}{0}}{blue}{0}}{black}{\attMid} \att{\att{\att{\att{CK20}{red}{\attHigh}}{green}{0}}{blue}{0}}{black}{\attHigh} \att{\att{\att{\att{-}{red}{\attHigh}}{green}{0}}{blue}{0}}{black}{0} \att{\att{\att{\att{,}{red}{\attHigh}}{green}{0}}{blue}{0}}{black}{\attMid} \att{\att{\att{\att{TTF}{red}{\attHigh}}{green}{0}}{blue}{0}}{black}{\attHigh} \att{\att{\att{\att{-}{red}{\attHigh}}{green}{0}}{blue}{0}}{black}{0} \att{\att{\att{\att{1}{red}{0}}{green}{0}}{blue}{0}}{black}{0} \att{\att{\att{\att{-}{red}{\attMid}}{green}{0}}{blue}{0}}{black}{0} \att{\att{\att{\att{,}{red}{\attMid}}{green}{0}}{blue}{0}}{black}{0} \att{\att{\att{\att{PROTEINA}{red}{\attHigh}}{green}{0}}{blue}{0}}{black}{\attMid} \att{\att{\att{\att{S}{red}{0}}{green}{0}}{blue}{0}}{black}{0} \att{\att{\att{\att{-}{red}{0}}{green}{0}}{blue}{0}}{black}{0} \att{\att{\att{\att{100}{red}{0}}{green}{0}}{blue}{0}}{black}{0} \att{\att{\att{\att{-}{red}{0}}{green}{0}}{blue}{0}}{black}{0} \att{\att{\att{\att{.}{red}{0}}{green}{0}}{blue}{0}}{black}{0} \att{\att{\att{\att{LESIONE}{red}{0}}{green}{0}}{blue}{0}}{black}{0} \att{\att{\att{\att{DI}{red}{0}}{green}{0}}{blue}{0}}{black}{0} \att{\att{\att{\att{CM}{red}{0}}{green}{0}}{blue}{0}}{black}{0} \att{\att{\att{\att{2}{red}{0}}{green}{0}}{blue}{0}}{black}{0} \att{\att{\att{\att{,}{red}{0}}{green}{0}}{blue}{0}}{black}{0} \att{\att{\att{\att{0}{red}{\attLow}}{green}{0}}{blue}{0}}{black}{0} \att{\att{\att{\att{X}{red}{0}}{green}{0}}{blue}{0}}{black}{0} \att{\att{\att{\att{1}{red}{0}}{green}{0}}{blue}{0}}{black}{0} \att{\att{\att{\att{,}{red}{0}}{green}{0}}{blue}{0}}{black}{0} \att{\att{\att{\att{3}{red}{0}}{green}{0}}{blue}{0}}{black}{0} \att{\att{\att{\att{X}{red}{0}}{green}{0}}{blue}{0}}{black}{0} \att{\att{\att{\att{0}{red}{0}}{green}{0}}{blue}{0}}{black}{0} \att{\att{\att{\att{,}{red}{0}}{green}{0}}{blue}{0}}{black}{0} \att{\att{\att{\att{7}{red}{0}}{green}{0}}{blue}{0}}{black}{0} \att{\att{\att{\att{.}{red}{0}}{green}{0}}{blue}{0}}{black}{0} \att{\att{\att{\att{1}{red}{0}}{green}{0}}{blue}{0}}{black}{0} \att{\att{\att{\att{-}{red}{0}}{green}{0}}{blue}{0}}{black}{0} \att{\att{\att{\att{2}{red}{0}}{green}{0}}{blue}{0}}{black}{0} \att{\att{\att{\att{)}{red}{0}}{green}{0}}{blue}{0}}{black}{0} \att{\att{\att{\att{SEZIONI}{red}{\attLow}}{green}{0}}{blue}{0}}{black}{0} \att{\att{\att{\att{SERIATE}{red}{\attHigh}}{green}{0}}{blue}{0}}{black}{0} \att{\att{\att{\att{.}{red}{\attMid}}{green}{0}}{blue}{0}}{black}{0}\end{minipage}&\begin{minipage}{\attTableTextWidth}\att{\att{\att{\att{LEFT}{red}{\attLow}}{green}{0}}{blue}{0}}{black}{0} \att{\att{\att{\att{PLEURAL}{red}{\attHigh}}{green}{\attMid}}{blue}{0}}{black}{0} \att{\att{\att{\att{EFFUSION}{red}{0}}{green}{0}}{blue}{0}}{black}{0} \att{\att{\att{\att{OF}{red}{0}}{green}{0}}{blue}{0}}{black}{0} \att{\att{\att{\att{UNKNOWN}{red}{0}}{green}{0}}{blue}{0}}{black}{0} \att{\att{\att{\att{ORIGIN}{red}{0}}{green}{0}}{blue}{0}}{black}{0} \att{\att{\att{\att{AND}{red}{0}}{green}{0}}{blue}{0}}{black}{0} \att{\att{\att{\att{LUNG}{red}{\attHigh}}{green}{0}}{blue}{0}}{black}{0} \att{\att{\att{\att{THICKENING}{red}{\attLow}}{green}{0}}{blue}{0}}{black}{0} \att{\att{\att{\att{OF}{red}{0}}{green}{0}}{blue}{0}}{black}{0} \att{\att{\att{\att{UNKNOWN}{red}{0}}{green}{0}}{blue}{0}}{black}{0} \att{\att{\att{\att{ORIGIN}{red}{0}}{green}{0}}{blue}{0}}{black}{0} \att{\att{\att{\att{,}{red}{0}}{green}{0}}{blue}{0}}{black}{0} \att{\att{\att{\att{ABDOMINAL}{red}{0}}{green}{0}}{blue}{0}}{black}{0} \att{\att{\att{\att{WALL}{red}{0}}{green}{0}}{blue}{\attLow}}{black}{0} \att{\att{\att{\att{NODULES}{red}{0}}{green}{0}}{blue}{0}}{black}{0} \att{\att{\att{\att{.}{red}{0}}{green}{0}}{blue}{0}}{black}{0} \att{\att{\att{\att{CANCEROUS}{red}{\attHigh}}{green}{0}}{blue}{0}}{black}{\attLow} \att{\att{\att{\att{INFILTRATION}{red}{0}}{green}{0}}{blue}{0}}{black}{0} \att{\att{\att{\att{OF}{red}{\attHigh}}{green}{0}}{blue}{0}}{black}{\attHigh} \att{\att{\att{\att{THE}{red}{\attHigh}}{green}{0}}{blue}{0}}{black}{\attHigh} \att{\att{\att{\att{CONNECTIVE}{red}{0}}{green}{0}}{blue}{0}}{black}{0} \att{\att{\att{\att{-}{red}{0}}{green}{0}}{blue}{0}}{black}{0} \att{\att{\att{\att{ADIPOSE}{red}{0}}{green}{0}}{blue}{0}}{black}{\attMid} \att{\att{\att{\att{STROMA}{red}{0}}{green}{0}}{blue}{0}}{black}{\attHigh} \att{\att{\att{\att{.}{red}{\attLow}}{green}{0}}{blue}{0}}{black}{0} \att{\att{\att{\att{IMMUNOHISTOCHEMICAL}{red}{\attHigh}}{green}{0}}{blue}{0}}{black}{\attHigh} \att{\att{\att{\att{:}{red}{\attLow}}{green}{0}}{blue}{0}}{black}{\attHigh} \att{\att{\att{\att{CK7}{red}{\attHigh}}{green}{0}}{blue}{0}}{black}{\attHigh} \att{\att{\att{\att{+}{red}{\attHigh}}{green}{0}}{blue}{0}}{black}{0} \att{\att{\att{\att{,}{red}{\attMid}}{green}{0}}{blue}{0}}{black}{\attMid} \att{\att{\att{\att{CK20}{red}{\attHigh}}{green}{0}}{blue}{0}}{black}{\attHigh} \att{\att{\att{\att{-}{red}{\attHigh}}{green}{0}}{blue}{0}}{black}{0} \att{\att{\att{\att{,}{red}{\attHigh}}{green}{0}}{blue}{0}}{black}{\attMid} \att{\att{\att{\att{TTF}{red}{\attHigh}}{green}{0}}{blue}{0}}{black}{\attHigh} \att{\att{\att{\att{-}{red}{\attHigh}}{green}{0}}{blue}{0}}{black}{0} \att{\att{\att{\att{1}{red}{0}}{green}{0}}{blue}{0}}{black}{0} \att{\att{\att{\att{-}{red}{\attMid}}{green}{0}}{blue}{0}}{black}{0} \att{\att{\att{\att{,}{red}{\attMid}}{green}{0}}{blue}{0}}{black}{0} \att{\att{\att{\att{PROTEIN}{red}{\attHigh}}{green}{0}}{blue}{0}}{black}{\attMid} \att{\att{\att{\att{S}{red}{0}}{green}{0}}{blue}{0}}{black}{0} \att{\att{\att{\att{-}{red}{0}}{green}{0}}{blue}{0}}{black}{0} \att{\att{\att{\att{100}{red}{0}}{green}{0}}{blue}{0}}{black}{0} \att{\att{\att{\att{-}{red}{0}}{green}{0}}{blue}{0}}{black}{0} \att{\att{\att{\att{.}{red}{0}}{green}{0}}{blue}{0}}{black}{0} \att{\att{\att{\att{2}{red}{0}}{green}{0}}{blue}{0}}{black}{0} \att{\att{\att{\att{CM}{red}{0}}{green}{0}}{blue}{0}}{black}{0} \att{\att{\att{\att{LESION}{red}{0}}{green}{0}}{blue}{0}}{black}{0} \att{\att{\att{\att{,}{red}{0}}{green}{0}}{blue}{0}}{black}{0} \att{\att{\att{\att{0}{red}{\attLow}}{green}{0}}{blue}{0}}{black}{0} \att{\att{\att{\att{X}{red}{0}}{green}{0}}{blue}{0}}{black}{0} \att{\att{\att{\att{1}{red}{0}}{green}{0}}{blue}{0}}{black}{0} \att{\att{\att{\att{,}{red}{0}}{green}{0}}{blue}{0}}{black}{0} \att{\att{\att{\att{3}{red}{0}}{green}{0}}{blue}{0}}{black}{0} \att{\att{\att{\att{X}{red}{0}}{green}{0}}{blue}{0}}{black}{0} \att{\att{\att{\att{0}{red}{0}}{green}{0}}{blue}{0}}{black}{0} \att{\att{\att{\att{,}{red}{0}}{green}{0}}{blue}{0}}{black}{0} \att{\att{\att{\att{7}{red}{0}}{green}{0}}{blue}{0}}{black}{0} \att{\att{\att{\att{.}{red}{0}}{green}{0}}{blue}{0}}{black}{0} \att{\att{\att{\att{1}{red}{0}}{green}{0}}{blue}{0}}{black}{0} \att{\att{\att{\att{-}{red}{0}}{green}{0}}{blue}{0}}{black}{0} \att{\att{\att{\att{2}{red}{0}}{green}{0}}{blue}{0}}{black}{0} \att{\att{\att{\att{)}{red}{0}}{green}{0}}{blue}{0}}{black}{0} \att{\att{\att{\att{SERIAL}{red}{\attHigh}}{green}{0}}{blue}{0}}{black}{0} \att{\att{\att{\att{SECTIONS}{red}{\attLow}}{green}{0}}{blue}{0}}{black}{0} \att{\att{\att{\att{.}{red}{\attMid}}{green}{0}}{blue}{0}}{black}{0}\end{minipage} \\
    \hline
  \end{tabular}
  \caption{Three sample reports annotated by the interpretable model (underline intensity proportional to class importance).}
  \label{tab:multiAttention1}
\end{figure*}
In the case of topography, when focusing on the performance on classes
with many examples, all models tend to perform similarly, with even
the interpretable model attaining high F1 scores. The advantage of
recurrent networks over bag-of-word representations is more pronounced
when focusing on rare classes. One possible explanation is that the
representation learned by recurrent networks is shared across all
classes, leveraging the advantage of multi-task
learning~\cite{caruana1997multitask} in this case.  We also note that
in no case hierarchical attention models outperform flat attention
models and max-pooling performs the best on rare classes. In the case
of morphology, differences among different models are more pronounced,
with \ac{bert} being very effective for densely populated classes (but
not for rare classes). Again hierarchical attention does not
outperform flat attention. This result differs from the ones reported
in~\cite{gao_hierarchical_2018} but the datasets are very different in
terms of number of examples and number of classes. Differences in the
writing style of pathologist trained and practicing in different
countries could also impact the relative performance of different
models. In this respect, our documents contain on average fewer
sentences (see \cref{fig:wordsDist} in Appendix~\ref{app:distr}),
offering less structure to be 
exploited by the richer hierarchical models.

The interpretable classifier \maxi{} can be used to explain prediction
by highlighting which portions of the text contribute to which
classes.  Its average agreement with \maxp{} was $91.8\%$ on
topography and $78.3\%$ on morphology.  In \cref{tab:multiAttention1},
we show three examples (topography task) where terms are underlined by
class-specific colors and with intensities proportional to the
importance $u_{j,t}$ of word in position $t$ for class $j$
(see \eqref{eq:maxiModelU}): high if $u_{j,t}>0.8$, medium if $u_{j,t}\in[0.3,0.8)$, low if
$u_{j,t}\in[0.1,0.3)$, not highlighted if $u_{j,t}<0.1$. We consider
class $j$ to be relevant to the document if at least one word has
$u_{j,t}\geq 0.1$.

The first report was correctly classified and the two most relevant
words are \emph{prostatico} (\emph{prostatic}), \emph{prostata}
(\emph{prostate}), followed by \emph{PSA} (Prostate-Specific Antigene)
and \emph{Gleason} score, that are two common exams in prostate cancer
cases \cite{brimo2013prostate}.
For the second report, the model proposes three codes: \emph{18}, \emph{20}
and \emph{21}, suggesting that \emph{intestinal tubular adenoma} and
\emph{pedunculated polypus} are terms associated with class colon,
\emph{polypus} associated with colon and rectum, and \emph{anal orifice}
associated with rectum and anus. Note that the ground truth for this record
was rectum, while the text explicitly mentions that the fragments have been
extracted at 20 cm from the anal orifice (the human rectum is
approximately 12 cm long and the anal canal 3-5 cm \cite{greene2006ajcc}).
The third report is an even more complex case
where the model proposes codes \emph{34}, attached to \emph{plurial effusion}
and \emph{lung thickening}, but interestingly also underlines the
immunohistochemical results, as the pattern \emph{CK7+} \emph{CK20-} commonly
indicates a diagnosis of lung origin for metastatic adenocarcinoma
\cite{kummar2002cytokeratin}. Also, immunohistochemistry is a common approach
in the diagnosis of tumors of uncertain origin
\cite{duraiyan2012applications}. This can be the reason for the underlying
with code \emph{80} of the immunoistochemical part.  It is interesting to note
that \emph{pleuric} is suggested to be related to ovarian cancer, in fact
the pleural cavity constitutes the most frequent site for extra-abdominal
metastasis in ovarian carcinoma \cite{porcel2012pleural}.

\begin{figure}
  \centering
  \includegraphics[width=0.45\textwidth]{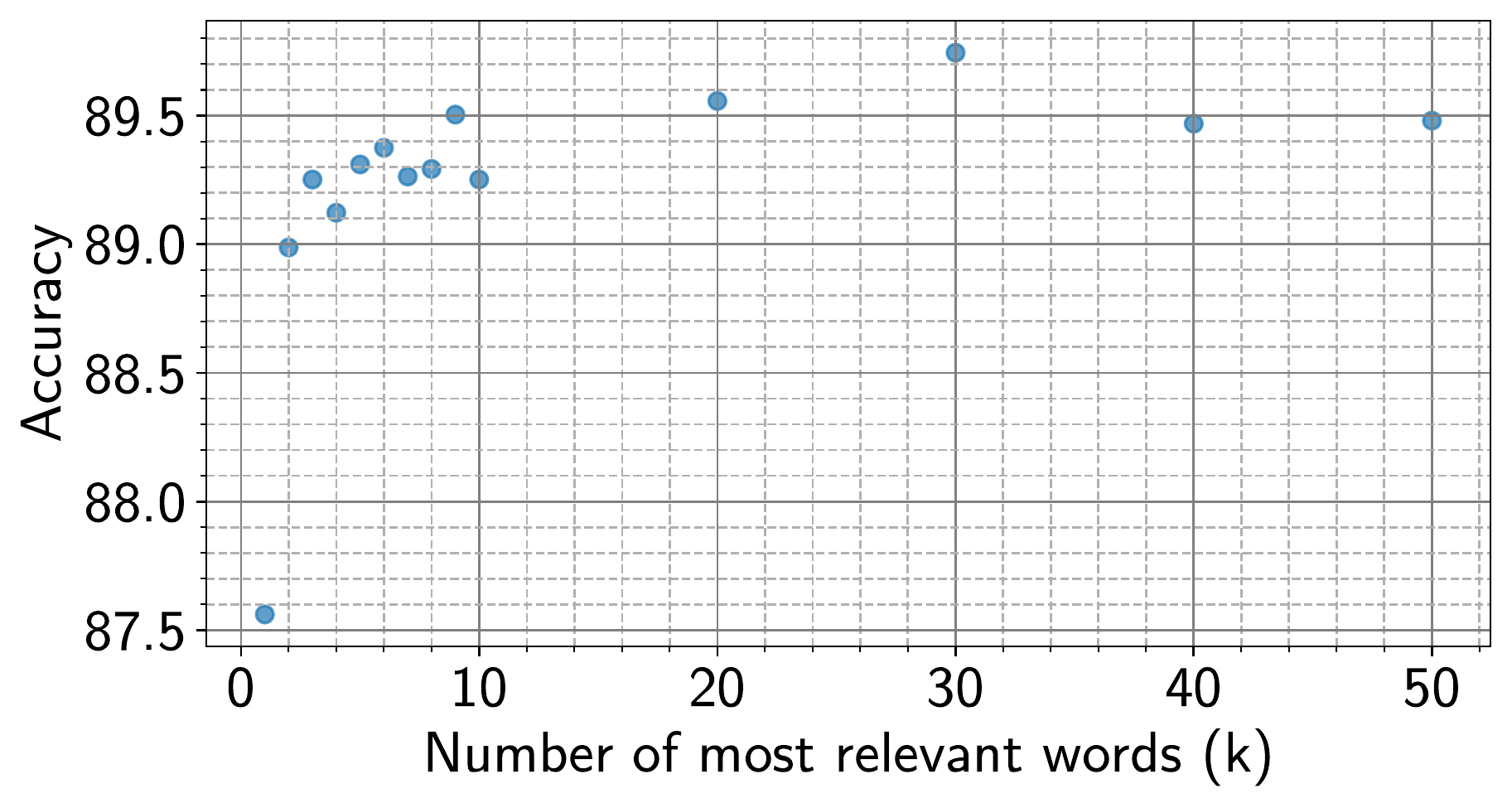}
  \caption{Training of a plain \ac{gru} model on topography task on
    datasets created 
    using \maxi{} to distill the most relevant $k$ words.}
  \label{fig:sintex}
\end{figure}
To quantify the effectiveness of the interpretable model, we designed an
experiment where a set of datasets is created taking only the most
relevant $k$ 
words based on the value of $u_{t}$ in \eqref{eq:maxiModelU} of
\maxi{} for the topography site prediction.
In~\cref{fig:sintex}, we plot the accuracy obtained training a plain \ac{gru}
model on those reduced datasets, for increasing values of $k$. Accuracy is high
even when selecting only a few words, suggesting that the interpretable model
is effective in distilling the most relevant terms, and that the information
contained in texts tends to be concentrated in a small number of terms.

\section{Conclusions}
We compared different algorithms on a large scale dataset of more than
$80\,000$ labeled records from the Tuscany region tumor registry
collected between 1990 and 2014.  Results confirm the viability of
automated assignment of \ac{icdo3} codes, with an accuracy of 90.3\% on
topography (61 classes) and 84.8\% on morphology (134 classes). Top-5
accuracies (fraction of test documents whose correct label is
among the top five model's prediction) were 98.1\% and 96.9\% for
topography and morphology, respectively.  The latter rates decreased
only to 96.2\% (topography) and 93.6\% (morphology) when using an
interpretable model that highlights the most important terms in the
text.

In this specific context we did not obtain significant improvements
using hierarchical attention methods, compared to a simple max pooling
aggregation. The difference between deep learning models and more
traditional approaches based on bag-of-words with SVM is significant
but not as pronounced as in the results reported in other
studies.
We also found that a large window size (15 words) and relatively small dimensionality (60) works better for construction of word vectors, while other
works in biomedical field~\cite{chiu2016train} found better results with smaller window size larger word vector
dimensionality.
These differences can be explained, at least in part, with
the specificity of the corpus used in this study, where reports tend
to be short, synthetic, rich in discriminant keywords, and often
lacking verb phrases. As shown in~\cref{fig:sintex}, few
words are sufficient to achieve good accuracy.

SVM perform well on topography class that are sufficiently well
represented in the dataset. 
Also, we found that hierarchical models are not
better than flat models and that a simple max aggregation achieves the
best results in most cases.  Interestingly, hierarchical models are
outperformed by flat attention or flat max pooling for the more
difficult classes (those with less than 100 training examples). Rare
classes remain however challenging for all current methods and as
discussed in~\cref{sec:dataset} our study, like all previous similar
studies in this area, do not even consider extremely rare
classes.
In this respect, future work may consider
the use of metalearning techniques capable of operating in the
few-shot learning setting~\cite{snell2017prototypical,%
  ravi2016optimization,vinyals2016matching} in order to include more
classes and to improve prediction accuracy on the underrepresented
ones. Results in this study are limited to a specific (but large) Italian dataset and might be compared in the future against results obtained on cancer reports written in other languages.

\bibliographystyle{ieeetr}
\bibliography{bibliography}

\appendices
\section{Dataset statistics}
\label{app:distr}
\begin{figure*}
  \centering
  \begin{tabular}{cc}
  \includegraphics[width=.38\textwidth]{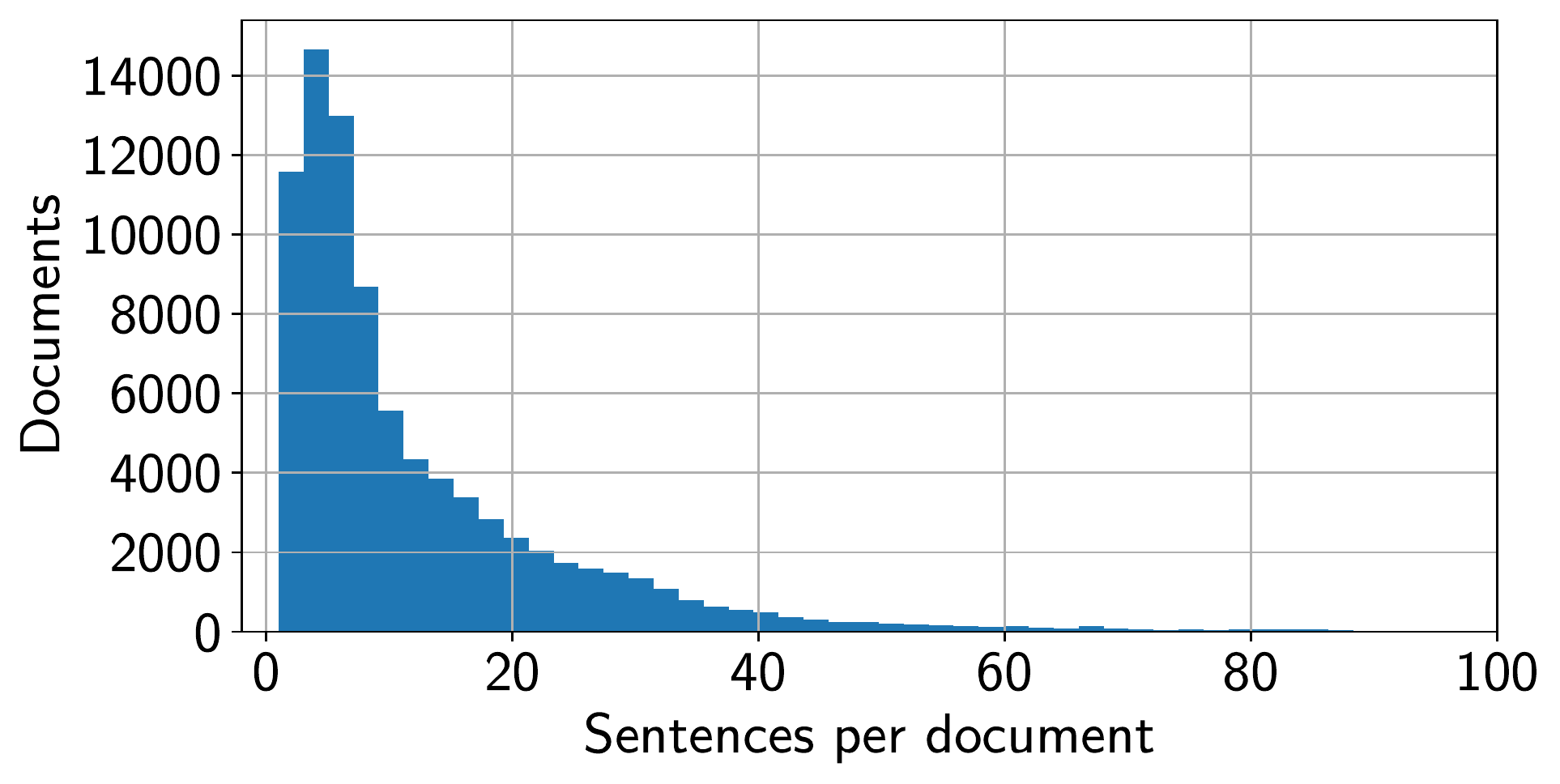}&
  \includegraphics[width=.38\textwidth]{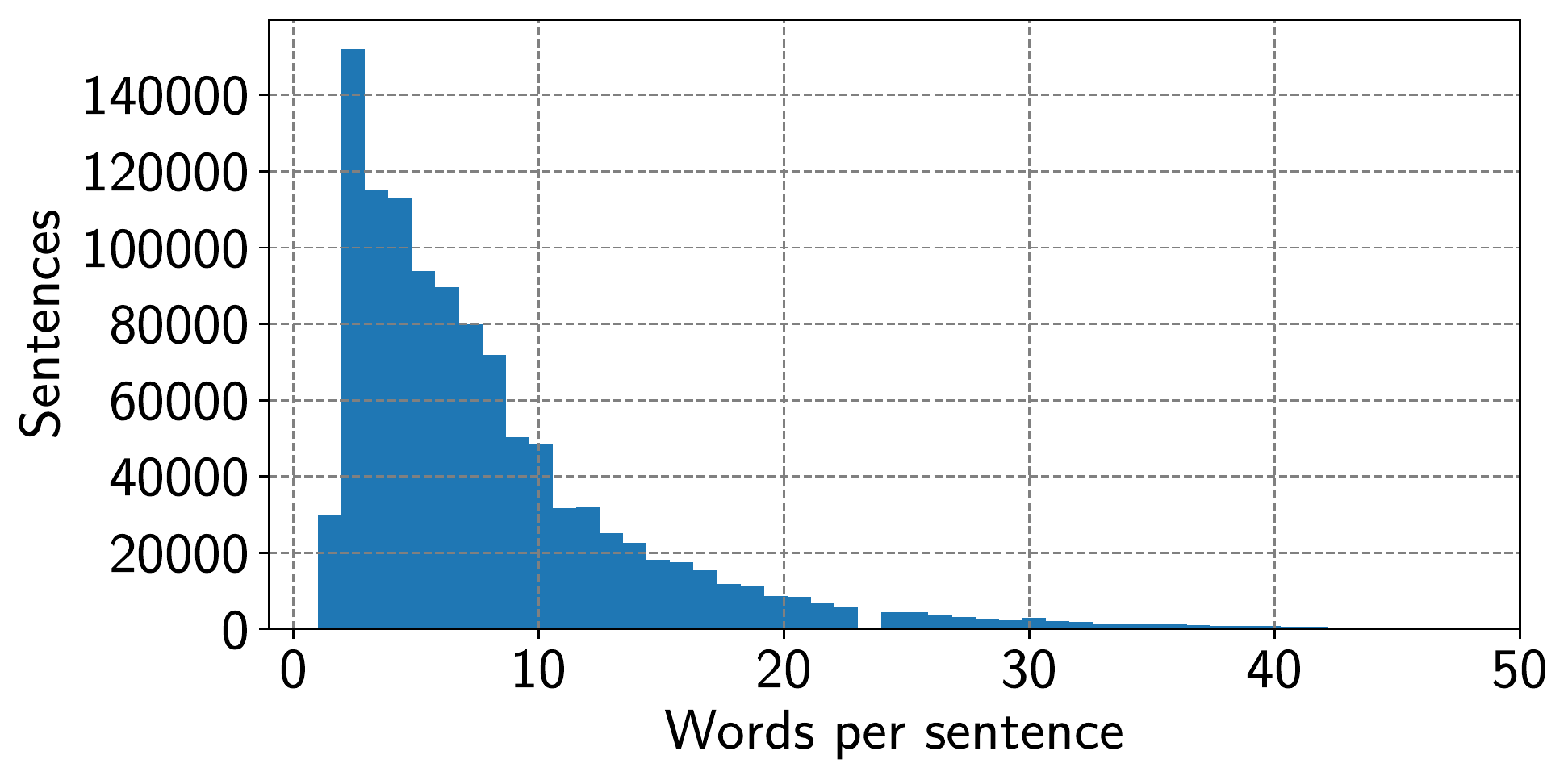}
  \end{tabular}
  \caption{Distribution of the number of sentences per document (left)
    and the number of words per sentence (right).}
  \label{fig:wordsDist}
\end{figure*}

\begin{figure*}
  \centering
  \begin{tabular}{cc}
  \includegraphics[width=.38\textwidth]{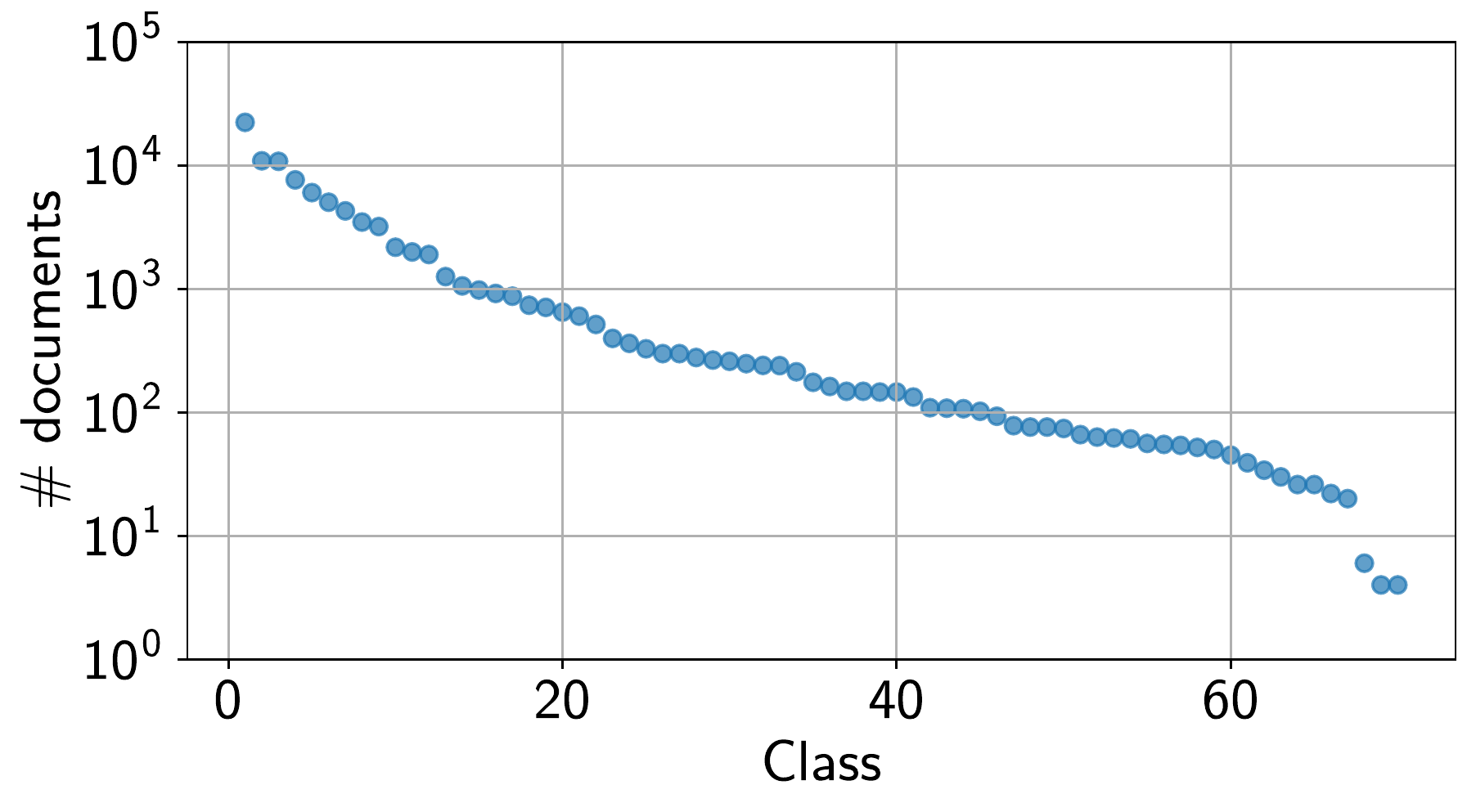}&
  \includegraphics[width=.38\textwidth]{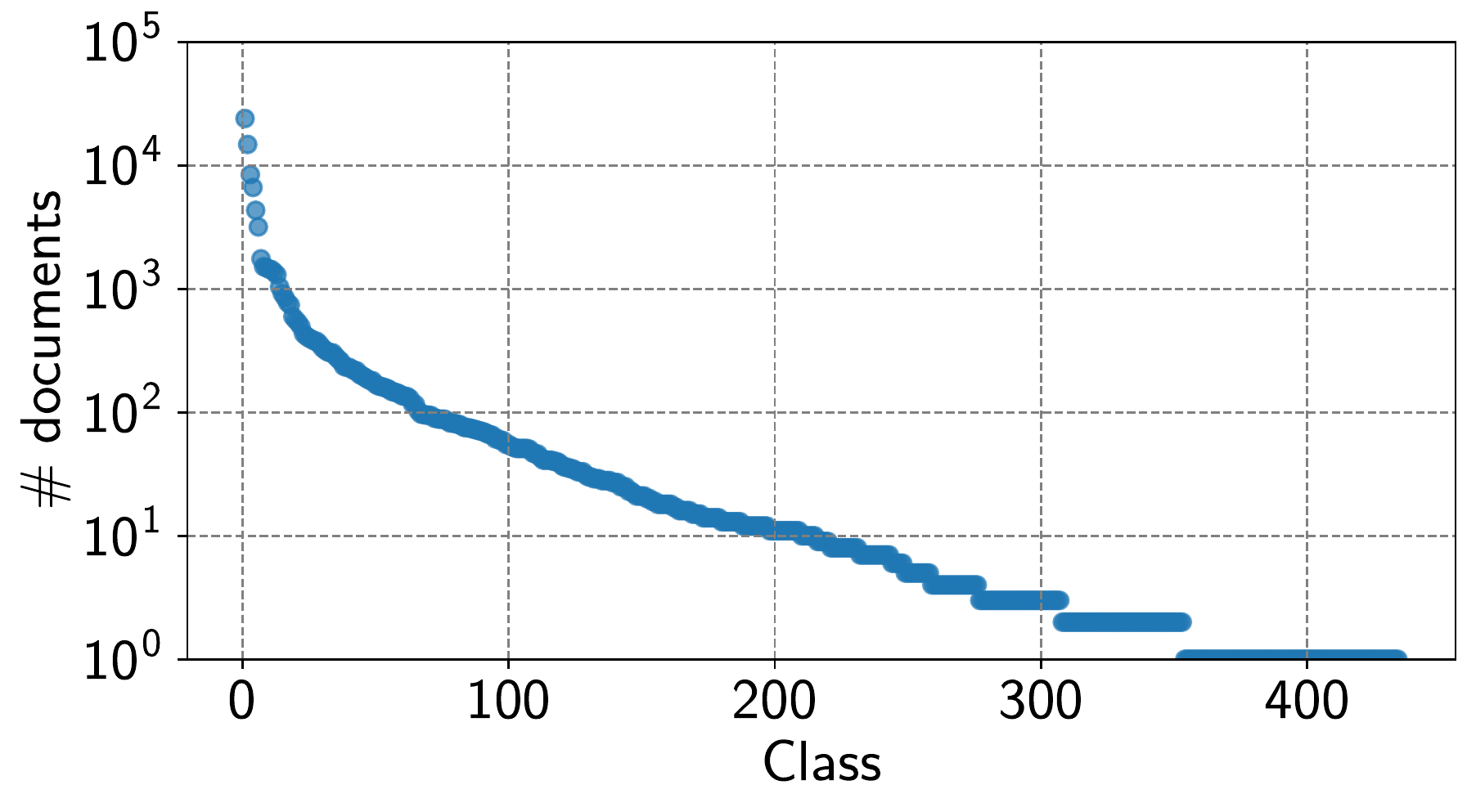}
  \end{tabular}
  \caption{Class distributions for topography (left) and morphology (right)}
  \label{fig:classDist}
\end{figure*}

We report in Figure~\ref{fig:wordsDist} and in Figure~\ref{fig:classDist} some
distributions of the dataset used in this study.

\section{Hyperparameter optimization}
\label{app:hyper}
We report here domains and optimal values (underlined) for the
hyperparameters of the models used in our experiments.

In \maxp{} we used the \emph{max} aggregation function in the plain
model of \cref{sec:modelh}. The hyperparameters
space was:
\begin{align*}
  \xi_{(l)}^f=\xi_{(l)}^r&\in[\underline{1},2],\qquad\xi_{(l)}^h\in[\underline{1},2,4],\\
  \xi_{(d)}^f=\xi_{(d)}^r&\in[2,4,8,16,32,64,\underline{128},256,512],\\
  \xi_{(d)}^h&\in[2,4,8,16,32,64,128,256,\underline{512},1024,2048],
\end{align*}
for the \emph{topography} site task, and:
\begin{align*}
  \xi_{(l)}^f=\xi_{(l)}^r&\in[\underline{1}],\qquad\xi_{(l)}^h\in[\underline{1},2,4],\\
  \xi_{(d)}^f=\xi_{(d)}^r&\in[2,4,8,16,32,64,\underline{128},256,512],\\
  \xi_{(d)}^h&\in[2,4,8,16,32,64,\underline{128},256,512,1024,2048],
\end{align*}
for the \emph{morphology} type task.

In \softmax{} we used the \emph{attention} aggregation function in the
plain model. The hyperparameters
space was:
\begin{align*}
  \xi_{(l)}^f=\xi_{(l)}^r&\in[\underline{1}],&\xi_{(l)}^h&\in[0,\underline{1}],\\
  \xi_{(d)}^f=\xi_{(d)}^r&\in[64,\underline{128},256],&\xi_{(d)}^h&\in[256,\underline{512},1024],\\
  \xi_{(d)}^a&\in[128,\underline{256},512,1024],
\end{align*}
for the site, and:
\begin{align*}
  \xi_{(l)}^f=\xi_{(l)}^r&\in[\underline{1}],&  \xi_{(l)}^h&\in[0,\underline{1}],\\
  \xi_{(d)}^f=\xi_{(d)}^r&\in[64,128,\underline{256}],&  \xi_{(d)}^h&\in[64,\underline{128},256],\\
  \xi_{(d)}^a&\in[128,\underline{256},512,1024],
\end{align*}
for the morphology.

In \maxh{} we used the \emph{max} aggregation in the hierarchical
model of \cref{sec:modelh}. The hyperparameters space was:
\begin{align*}
  \xi_{(l)}^f=\xi_{(l)}^r=\bar{\xi}^f_{(l)}=\bar{\xi}^r_{(l)}&\in[\underline{1}],\qquad\xi_{(l)}^h\in[0,1,\underline{2},4],\\
  \xi_{(d)}^f=\xi_{(d)}^r=\bar{\xi}^f_{(d)}=\bar{\xi}^r_{(d)}&\in[32,\underline{64},128,256],\\
  \xi_{(d)}^h&\in[256,512,\underline{1024},2048],\\
\end{align*}
for the topography, and:
\begin{align*}
  \xi_{(l)}^f=\xi_{(l)}^r=\bar{\xi}^f_{(l)}=\bar{\xi}^r_{(l)}&\in[\underline{1}],\qquad\xi_{(l)}^h\in[0,\underline{1},2,4],\\
  \xi_{(d)}^f=\xi_{(d)}^r=\bar{\xi}^f_{(d)}=\bar{\xi}^r_{(d)}&\in[32,\underline{64},128,256],\\
  \xi_{(d)}^h&\in[256,512,\underline{1024},2048],\\
\end{align*}
for the morphology.

In \softmaxh{} we used the \emph{attention} aggregation in the
hierarchical model. The hyperparameters space was:
\begin{align*}
  \xi_{(l)}^f=\xi_{(l)}^r=\bar{\xi}^f_{(l)}=\bar{\xi}^r_{(l)}&\in[\underline{1}],\qquad\xi_{(l)}^h\in[0,\underline{1},2,4],\\
  \xi_{(d)}^f=\xi_{(d)}^r=\bar{\xi}^f_{(d)}=\bar{\xi}^r_{(d)}&\in[32,64,\underline{128},256],\\
  \xi_{(d)}^h&\in[256,512,\underline{1024},2048],\\
  \xi_{(d)}^a=\bar{\xi}^a_{(d)}&\in[64,\underline{128},256,512],
\end{align*}
for the topography, and:
\begin{align*}
  \xi_{(l)}^f=\xi_{(l)}^r=\bar{\xi}^f_{(l)}=\bar{\xi}^r_{(l)}&\in[\underline{1}],\qquad\xi_{(l)}^h\in[0,\underline{1},2,4],\\
  \xi_{(d)}^f=\xi_{(d)}^r=\bar{\xi}^f_{(d)}=\bar{\xi}^r_{(d)}&\in[32,\underline{64},128,256],\\
  \xi_{(d)}^h&\in[256,512,\underline{1024},2048],\\
  \xi_{(d)}^a=\bar{\xi}^a_{(d)}&\in[64,\underline{128},256,512],
\end{align*}
for the morphology.

In \maxi{} we used the \emph{max} aggregation in the plain
model. Also we set the model to be interpretable. The hyperparameters
space was: 
\begin{align*}
  \xi_{(l)}^f=\xi_{(l)}^r&\in[1,\underline{2},4],\qquad\xi_{(l)}^h\in[\underline{1},2,4],\\
  \xi_{(d)}^f=\xi_{(d)}^r&\in[2,4,8,16,32,64,\underline{128},256,512],\\
  \xi_{(d)}^h&\in[2,4,8,16,32,64,128,256,512,1024,2048],
\end{align*}
for the topography, and:
\begin{align*}
  \xi_{(l)}^f=\xi_{(l)}^r&\in[1,\underline{2},4],&\xi_{(l)}^h&\in[\underline{1}],\\
  \xi_{(d)}^f=\xi_{(d)}^r&\in[64,128,\underline{256},512],&\xi_{(d)}^h&\in[],
\end{align*}
for the morphology. Note that, in this setting, the size of the
last layer of $G$ must be equal to the output size of the model
(and the softmax is applied directly after the aggregation $A$,
without any layer). Thus, $\xi_{(d)}^h$ refers only to the
layers before the last one, if they exist.

Regarding \gru{}, we searched in a space of $[1,2,4]$ number of layers of
dimension in $[128,256,512,1024]$. We found that the best
configuration was using $2$ layers of dimension $256$.

\section{Performance measures}
\label{app:measures}
We report in the following precise definitions of our performance measures.
\begin{itemize}
\item The multiclass accuracy is defined as
  $$
  A\doteq\frac{1}{m}\sum_{i=1}^m \ONE{y^{(i)} = \argmax_{j=1,\dots,K} f_j(x^{(i)})},
  $$
  where $\ONE{}$ denotes the indicator function and $m$ is the number of test
  points (recall that $f(x)$ denotes the vector of conditional probabilities
  assigned to each of the $K$ classes). It is equivalent to micro-averaged F1
  measure for mutually exclusive classes.
\item The top-$\ell$ accuracy is defined as
  $$
  A_\ell\doteq\frac{1}{m}\sum_{i=1}^m \ONE{y^{(i)} \in T_\ell\left(f(x^{(i)})\right)},
  $$
  where $T_\ell(a)$ denotes the operator that given array $a=[a_1,\dots,a_K]$
  as input returns the set $\{\pi_1,\dots,\pi_\ell\}$ being
  $\pi_1,\dots,\pi_K$ the permutation sequence that sorts $a$ in descending
  order
\item The macro-averaged F1 measure is defined as
  $$
  F_1^M \doteq \frac{1}{K}\sum_{k=1}^K \frac{2P_kR_k}{P_k+R_k}
  $$
  where
  $$
  P_k=\frac{\displaystyle\sum_{i=1}^m \ONE{y^{(i)}=k} \ONE{ y^{(i)}=\argmax_{j=1,\dots,K} f_j(x^{(i)})}}
  {\displaystyle\sum_{i=1}^m \ONE{k=\argmax_{j=1,\dots,K} f_j(x^{(i)})}}
  $$ is the precision for class $k$ and
  $$
  R_k=\frac{\displaystyle\sum_{i=1}^m \ONE{y^{(i)}=k} \ONE{ y^{(i)}=\argmax_{j=1,\dots,K} f_j(x^{(i)})}}
  {\displaystyle\sum_{i=1}^m \ONE{ k=y^{(i)}}}
  $$ is the recall for class $k$;
\item The fidelity is defined as
  $$
  F\doteq\frac{1}{m}\sum_{i=1}^m \ONE{\argmax_{j=1,\dots,K} f_j(x^{(i)}) = \argmax_{j=1,\dots,K} g_j(x^{(i)})},
  $$
  where $f(x)$ and $g(x)$ denote the vectors of conditional probabilities
  assigned to each of the $K$ classes by the two models (\maxp{} and
  \maxi{} in the paper).
\end{itemize}
\end{document}